\pdfoutput=1
\def\year{2022}\relax
\documentclass[letterpaper]{article} 
\usepackage{aaai22}  
\usepackage{times}  
\usepackage{helvet}  
\usepackage{courier}  
\usepackage[hyphens]{url}  
\usepackage{graphicx} 
\usepackage{natbib}  
\usepackage{caption} 
\DeclareCaptionStyle{ruled}{labelfont=normalfont,labelsep=colon,strut=off} 
\usepackage{booktabs}
\usepackage{newfloat}
\usepackage{listings}
\usepackage{amsmath}
\usepackage{amssymb}
\usepackage{multicol}
\usepackage[ruled,linesnumbered]{algorithm2e}
\usepackage{float}

\title{Neural Marionette: 
Unsupervised Learning of Motion Skeleton \\and Latent Dynamics from Volumetric Video}

\author {
    Jinseok Bae,
    Hojun Jang, 
    Cheol-Hui Min,
    Hyungun Choi,
    Young Min Kim
}

\affiliations {
    Seoul National University\\
    \{capoo95, j12040208, mch5048, woody101, youngmin.kim\}@snu.ac.kr
}

\begin{document}

\maketitle

\begin{abstract}
We present Neural Marionette, an unsupervised approach that discovers the skeletal structure from a dynamic sequence and learns to generate diverse motions that are consistent with the observed motion dynamics.
Given a video stream of point cloud observation of an articulated body under arbitrary motion, our approach discovers the unknown low-dimensional skeletal relationship that can effectively represent the movement.
Then the discovered structure is utilized to encode the motion priors of dynamic sequences in a latent structure, which can be decoded to the relative joint rotations to represent the full skeletal motion.
Our approach works without any prior knowledge of the underlying motion or skeletal structure, and we demonstrate that the discovered structure is even comparable to the hand-labeled ground truth skeleton in representing a 4D sequence of motion.
The skeletal structure embeds the general semantics of possible motion space that can generate motions for diverse scenarios.
We verify that the learned motion prior is generalizable to the multi-modal sequence generation, interpolation of two poses, and motion retargeting to a different skeletal structure.
\end{abstract}

\section{1\quad Introduction}

The skeletal structure of an articulated body~\cite{ceccarelli2004international} has been widely deployed for robotics control~\cite{veerapaneni2020entity, ha2020learning} or character animations~\cite{xu2019predicting, liu2019neuroskinning, yang2020transmomo}.
The low-dimensional motion structure can act as an important cue to detect accurate movement and provide interaction between a human and an intelligent agent in a complex environment.
Successful applications usually rely on strong priors such as human body joints, hands, or faces~\cite{zuffi2015stitched, SMPL-X:2019, zimmermann2021contrastive, schmidtke2021unsupervised} incorporated with the recent deep learning architecture.
However, it is challenging to obtain the accurate structure of an unknown subject from raw observation.
Some works extract skeleton using geometric priors, such as medial axis transform~\cite{lin2021point2skeleton} or low-dimensional primitives~\cite{paschalidou2019superquadrics, paschalidou2021neural}, 
while others discover the unknown motion prior for 4D tracking or motion prediction in temporally dense observation~\cite{bozic2021neural, li20214dcomplete, lin2020improving}.
But they are not designed to understand the motion semantics that can cover a large variation of plausible motion of a subject with an unknown skeletal structure.

\begin{figure}[t]
\centering
\includegraphics[width=0.45\textwidth]{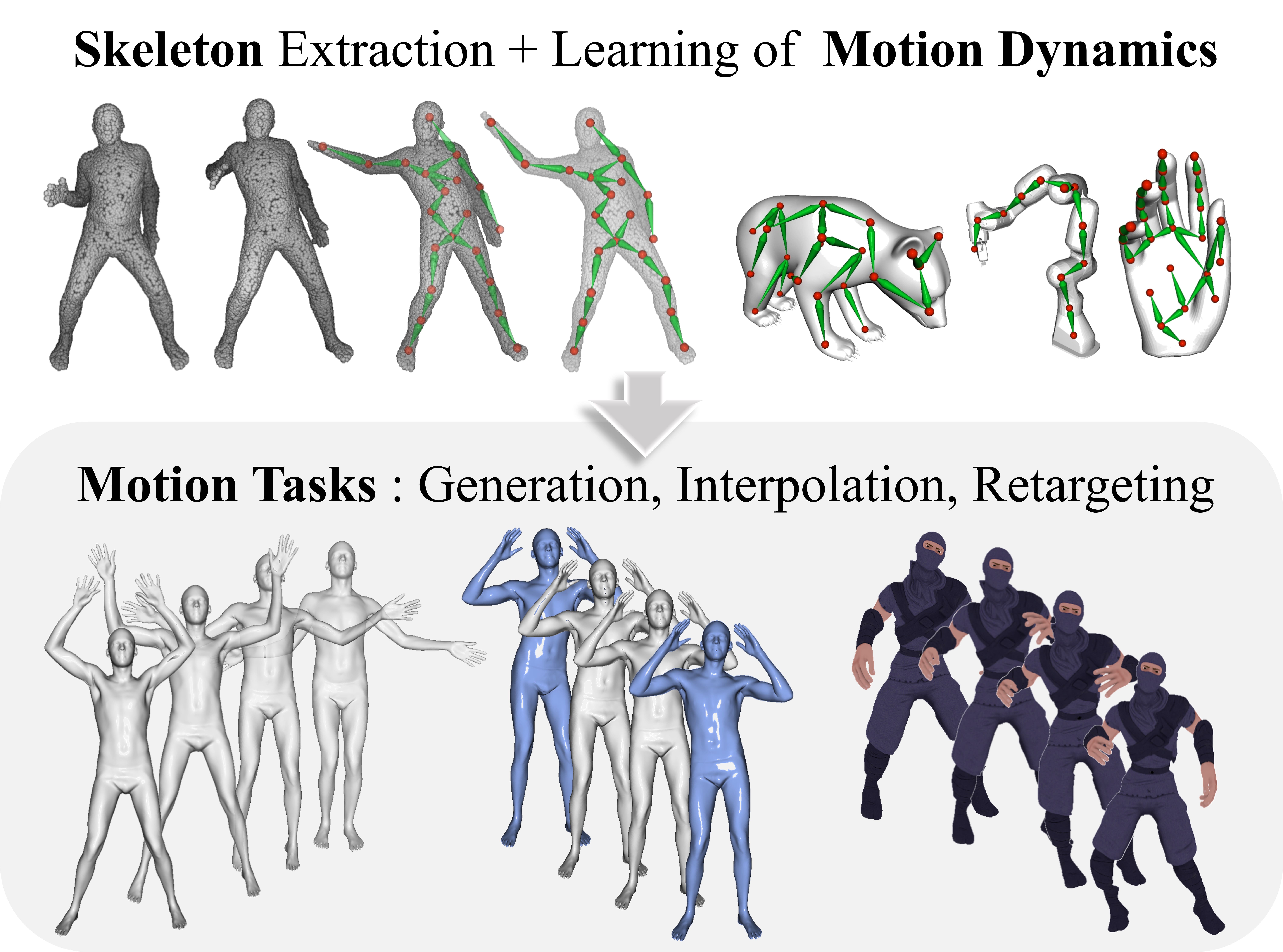} 
\caption{Overview of Neural Marionette. Our model learns to discover adaptive skeleton from volumetric video, and finds rules of motion by observing movements of skeletons.}
\label{fig:overview}
\end{figure}

In this paper, we propose Neural Marionette, a fully unsupervised framework that discovers a semantically consistent skeleton from a 3D motion sequence and learns the general motion dynamics of the discovered structure, which is illustrated in Figure~\ref{fig:overview}. 
Our framework consists of two main stages: a \emph{skeleton module} that determines explicit skeleton for motion, and a \emph{dynamics module} that learns to propagate the skeletons along the time axis.
Given point cloud sequence capturing the dynamic movement of an articulated body, our proposed model first learns to detect a skeleton tree without any prior knowledge of the topology.

To detect candidate nodes of a skeletal graph, we adapt the keypoint detectors, which have been demonstrated to be simple yet powerful in reasoning motions from video~\cite{minderer2019unsupervised,li2020causal,suwajanakorn2018discovery,chen2021unsupervised}.
Our work extends the unsupervised keypoint detection and explicitly models the parent-child relationship between the detected keypoints to build a skeletal tree, which serves as a powerful prior of structure to represent the motion dynamics in the subsequent stage.
Given the discovered topological graph of the skeleton, the dynamics module is formulated with recurrent neural networks to embed the sequence of motion into stochastic latent variables.
Specifically, the latent variable encodes the local rotation of each joint rather than location, so that general motion priors are effectively captured from different skeletons.
The embedding is completely task-agnostic, and can generate a sequence of skeletal motion for any downstream tasks without further adaptation.

We demonstrate that our model can extract skeletons of various topologies including full-body humans, robots, hands, and animals. Then we evaluate the performance of the skeletal structure in motion reconstruction.
Interestingly, we find that the structure discovered from our model sometimes outperforms the hand-labeled ground truth skeleton in 4D tracking.
The learned dynamics is verified to generate plausible motions for three different downstream tasks:  motion generation, interpolation, and retargeting. 
To the best of our knowledge, Neural Marionette is the first work to learn skeleton and latent dynamics from sequential 3D data, which does not exploit any categorical prior knowledge nor optimize for a specific sequence to enhance performance.

\section{2\quad Related Works}
\subsection{Understanding motion}
In this work, we focus on understanding the motion of the articulated body which could be represented in terms of the skeletal structure. Our approach jointly learns the motion structure (skeleton) and the possible movement (motion dynamics), and each has been investigated in the literature.
The motion structure is a shared topology of the skeleton to represent a given class of bodies.
If the target class is known, for example human bodies, parametric 3D models~\cite{anguelov2005scape, loper2015smpl, romero2017embodied, zuffi20173d} are acquired with a large amount of human annotations.
Parametric models exhibit successful achievement in a variety of applications such as shape reconstruction and pose estimation.
When the structure is unknown~\cite{palafox2021npms}, data-driven approaches can excavate structure from observations~\cite{xu2019unsupervised, xu2020rignet, lin2021point2skeleton} with self-supervised approaches that encourage consistent topology.
However, the inferred structure often is prone to errors and consequently suffers from performance degradation in motion analysis compared to sophisticated templates learned from labeled data.

After the skeletal structure defines the body as a combination of locally rigid parts, there exist a set of possible joint configurations of the given skeleton to perform plausible natural motion.
Given the topology of skeleton, recent studies utilize graph neural networks to learn the complex motion patterns~\cite{guo2019human, mao2019learning, liu2020disentangling}. 
However, they heavily rely on accurate skeleton, and the performance is usually demonstrated in human body or production characters.
To our knowledge, no previous works can discover the unknown skeletal structure and its movement that can accurately generate a large class of semantic motions.

\subsection{Variational recurrent models}

We train a generative model to represent a set of plausible motions of the given graphical structure.
Variational autoencoder (VAE)~\cite{kingma2013auto} builds a latent space of the data observation that follows the desired distribution and demonstrates promising results in various generative tasks of computer vision~\cite{eslami2016attend,crawford2019spatially, burgess2019monet, engelcke2019genesis}.

The latent representation can be further extended to include the temporal context of sequential data like video or speech by propagating hidden states through recurrent neural networks (RNN)~\cite{srivastava2015unsupervised}. 
Variational recurrent neural network (VRNN)~\cite{chung2015recurrent} is the recurrent version of VAE, which models the dependency of latent variables between neighboring timesteps. 
A number of works~\cite{kosiorek2018sequential, minderer2019unsupervised, hajiramezanali2019variational, veerapaneni2020entity, lin2020improving} demonstrated promising results of VRNN on tasks like video prediction and dynamic link prediction.
Our work also temporally extends the embedding of skeletal motion using VRNN and can successfully generate the motion sequence of the discovered skeleton.

\section{3\quad Background}
\label{sec:background}

\subsection{Forward kinematics}
Our skeletal structure defines the motion with the forward kinematics, which we introduce here.
The position of a $k$-th node for skeleton in a canonical pose $\mu_{c,k}$ provides offset $d_k\in\mathbb{R}^3$ from its parent
\begin{equation}
    d_k = \mu_{c,k} - \mu_{c, parent(k)}.
    \label{eq:offset}
\end{equation}
The length of the displacement $\|d_k\|$  represents the length of the bone connecting the $k$-th node and its parent, and is preserved under any possible deformation.
A new pose of the skeleton is composed of a global translation of the root node and a set of local rotations of each joint. 
Specifically, the chain of forward kinematics represents the  joint locations as
\begin{equation}
    \begin{gathered}
        \mu_k = \mu_{parent(k)} + R_kd_k \text{ where}\\
        R_k = \tilde{R}_{root} \cdots \tilde{R}_{parent(k)}\tilde{R}_k=R_{parent(k)}\tilde{R}_k.\\
    \end{gathered}
    \label{eq:fk}
\end{equation}
Here $\mu_k$ refers to the joint position in the current pose, and $\tilde{R}_k\in\mathbb{R}^{3\times3}$ refers to the relative rotation with respect to their parents in a local coordinate.
In summary, forward kinematics encodes a pose of a skeleton with a set of rotation matrices, once the skeletal structure defines the node positions at the rest pose and directed links of parent-child relationship.

\subsection{Variational recurrent neural network}
Variational recurrent neural network represents the observations of time steps $x_t$ with a VAE that is composed of a prior distribution $p_\theta(z_t|h_t)$  and a posterior distribution $q_\phi(z_t|x_t,h_t)$. $z_t$ is the latent variable that encodes the observation $x_t$, and sampled for generation. In addition to the ordinary VAE, both prior and posterior are conditioned on the hidden state of RNN $h_t$.
The latent variable $z_t$ is trained to maximize the likelihood of overall observation $x_t$ with reconstructed input $\hat{x}_t$ from the decoder $\varphi$ by minimizing
\begin{equation}
    \mathcal{L}_{vrnn}={1\over T}\sum_t \|x_t - \hat{x}_t\|_2^2,~\text{where}~\hat{x}_t = \varphi(z_t, h_t).
    \label{eq:vrnn_decoder}.
\end{equation}
At the same time, $p_\theta$ is encouraged to track $q_\phi$ with the KL divergence term of $\mathcal{L}_{kl}$ to learn the distribution that matches the observation with evidence lower bound (ELBO)~\cite{kingma2013auto}
\begin{equation}
        \mathcal{L}_{kl} = {1\over T}\sum_t D_{KL}(q_\phi(z_t|x_t,h_t)\|p_\theta(z_t|h_t)).\label{eq:loss_dynamics}
\end{equation}

\begin{figure}[t]
\centering
\includegraphics[width=0.45\textwidth]{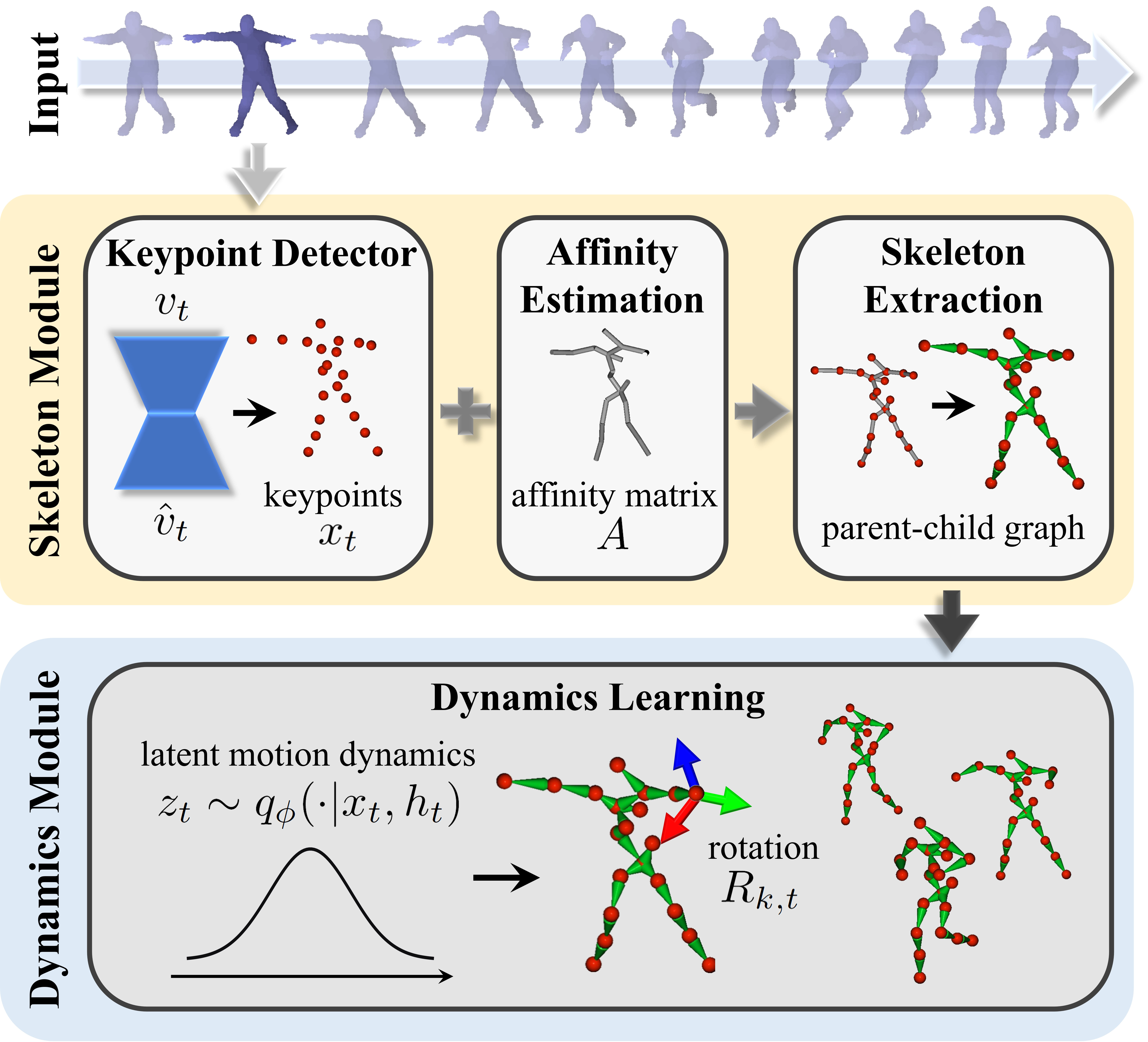} 
\caption{System pipeline of Neural Marionette. Given a voxelized sequence, skeleton module extracts skeleton, and dynamics module learns motion dynamics from the observed skeletal motion.}
\label{fig:pipeline}
\end{figure}

\section{4\quad Neural Marionette}

Neural Marionette  is composed of two stages: \emph{skeleton module} that extracts the underlying skeleton, and \emph{dynamics module} that learns the motion dynamics associated with the discovered skeleton.
The overall process is described in Figure~\ref{fig:pipeline}.
Detailed structures and learning strategies of the two modules are explained below.

\subsection{Skeleton module}

Given a sequence of point cloud observations, 
the unstructured point cloud is first discretized into binary voxels in a normalized grid  $v_t\in\mathbb{R}^{G_x\times G_y\times G_z}$, $t=1,\ldots,T$ such that we can efficiently access the neighboring observation.
The skeleton module extracts the frames of keypoints from the voxelized sequences and connects the neighboring keypoints to build a shared skeleton structure for the dynamics module.
The consistent skeleton acts as a light-weight and very efficient structural prior for extracting motion semantics of complex deformation.

\paragraph{Keypoint detector}

The keypoint detector is an encoder-decoder framework that maps the voxelized sequence $v_t$ into the trajectory of $K$ keypoints $\{\{x_{k,t}\}_K\}_T\in\mathbb{R}^{4 \times K \times T}$.
Each $x_{k,t}$ consists of a 3D location $\mu_{k,t}\in\mathbb{R}^3$ and intensity $\alpha_{k,t}\in\mathbb{R}$. 
Because the embedding represents the physical 3D coordinates of the keypoints, and the motion extracted with the keypoints becomes extremely interpretable in the subsequent dynamics module.

The keypoint extractor is inspired from autoencoder-based keypoint detectors on 2D video~\cite{jakab2018unsupervised, minderer2019unsupervised} and extended to 3D.
The encoder first extracts grid features $c_t \in \mathbb{R}^{D\times G'_x\times G'_y \times G'_z}$ with $D$ channels in a condensed resolution as
\begin{equation}
    c_t=f_{feat}(v_t),
    \label{eq:vox_enc_feature}
\end{equation}
and regresses $K$ heatmaps $m_t\in\mathbb{R}^{K\times G'_x\times G'_y \times G'_z}$
\begin{equation}
    m_t = f_{heat}(c_t, {1\over T}\sum_t{v_t}).
    \label{eq:vox_enc_heatmap}
\end{equation}
While the conventional keypoint detectors~\cite{jakab2018unsupervised, minderer2019unsupervised} do not explicitly share features between different time steps, our keypoint detector is augmented with the temporal mean of $v_{1:T}$ as shown in Eq.~\eqref{eq:vox_enc_heatmap}, from which the network can observe the spatio-temporal context.
Each channel in $m_t$ represents a probabilistic distribution of keypoint and we extract the keypoints $x_{k,t}=(\mu_{k,t}, \alpha_{k,t})$ from it.

The decoder is trained to recover the original sequence $v_t$ from the keypoints $x_t$ extracted from the encoder,
\begin{equation}
    \begin{gathered}
        \hat{v}_t =  f_{dec}(g_t,g_1,c_1,v_1),\\
        \text{where }g_{k,t}=\mathcal{N}_{grid}(\mu_{k,t}, \sigma_g)~~\forall{k}.
    \end{gathered}\label{eq:skel_dec}
\end{equation}
$\mathcal{N}_{grid}(\mu_{k,t}, \sigma_g)$ is Gaussian distribution discretized in a grid, with mean at $\mu_{k,t}$ and variance of hyper-parameter $\sigma_g$.
Basically, the keypoint is transformed into a synthetic heatmap $g_t$ by convolving the Gaussian kernel around the keypoint locations and reconstruct voxel difference $v_t - v_1$ with the decoding network $f_{dec}$.
Focusing on the difference encourages the keypoints to capture dynamic area, where $v_t - v_1$ is non-zero~\cite{minderer2019unsupervised}.

The encoder-decoder network is trained to find the optimal keypoints that best describes the motion of occupied voxels. 
Because the network encourages keypoints to capture the changing voxels, the keypoints might ignore static region.
We explicitly suggest to uniformly spread the keypoints within the point cloud $V_t$ that represents 3D coordinates $p_t$ of occupied voxels in $v_t$ with the volume fitting loss
\begin{equation}
        \mathcal{L}_{vol} = {1\over T}\sum_t{1\over|V_t|}\sum_{p_t\in V_t}\min_k\|p_t-\mu_{k,t}\|_2^2,\\
    \label{eq:volume_fitting_loss}
\end{equation}
that minimizes the one-directional Chamfer Distance~\cite{fan2017point} of $V_t$ from keypoints $x_t$.

The loss function to train the autoencoder includes additional terms from previous work, namely the reconstruction loss, sparsity loss, and the separation loss.
The reconstruction loss $\mathcal{L}_{recon}$ is the basic loss for an autoencoder, where we want to best reconstruct the original voxel,
\begin{equation}
    \mathcal{L}_{recon} = {1\over T}\sum_t\text{BCE}(v_t, \hat{v}_t).
    \label{eq:recon_loss}
\end{equation}
The proposed volume fitting loss complements the conventional reconstruction loss by capturing the static body parts.
The remaining two loss terms are adapted from the state-of-the-art keypoint detector~\cite{minderer2019unsupervised}.
The sparsity loss $\mathcal{L}_{sparse}$ enforces sparsity of the heatmap, 
\begin{equation}
    \mathcal{L}_{sparse} = {1\over TK}\sum_t\sum_k{\|m_{k,t}\|},
    \label{eq:sparsity_loss}
\end{equation}
and the separation loss $\mathcal{L}_{sep}$ encourages different trajectories  between keypoints
\begin{equation}
    \mathcal{L}_{sep} = {1\over {TK(K-1)}}\sum_t\sum_k\sum_{k'\neq k}e^{-\sigma_{s}\|s_{k,t}-s_{k',t}\|_2^2},
    \label{eq:separation_loss}
\end{equation}
where $s_{k,t}=\mu_{k,t}-{1\over T}\sum_t\mu_{k,t}$ and $\sigma_{s}$ is a hyper-parameter.

\paragraph{Affinity estimation}
Along with the learning of keypoints, our skeleton module estimates the affinity $A\in\mathbb{R}^{K\times K}$ between keypoints in order to compose edges of the skeleton. 
We first build decomposed affinity matrices $\{A_n\in\mathbb{R}^{K\times K}\}_N$ that focuses on $N(N<K)$  nearest neighbors of keypoints that can be combined to the final affinity matrix
\begin{equation}
        a_{ij} = \max_n{a_{n,ij}},~~a_{ij}\in A, a_{n,ij}\in A_n.
\label{eq:loss_kypt}
\end{equation}
Our affinity estimator builds on the prior work~\cite{bozic2021neural} that considers the position of the nodes as a strong prior on connections.
In addition to the previously suggested losses that observe a single frame, we propose the graph trajectory loss $\mathcal{L}_{traj}$ that encourages connectivity $a_{kk'}$ between the keypoints moving in the similar path 
\begin{equation}
    \mathcal{L}_{traj} = {1\over TK^2}
        \sum_t\sum_k\alpha_{k,t}\sum_{k'}a_{kk'}\text{C}(\mu_{k,t}, \mu_{ k',t})
        \label{eq:graph_traj_loss}
\end{equation}
given keypoint positions $\mu_{k,t}$ and intensities $\alpha_{k,t}$.
$\text{C}(\cdot)$ is a function that depends on the velocities $\dot{\mu}$ and accelerations $\ddot{\mu}$ of keypoints
\begin{equation}
        \text{C}(\mu_{k,t}, \mu_{k', t}) = {1\over2} - {1\over4}\left({{\langle\dot{\mu}_{k,t}, \dot{\mu}_{k',t}\rangle}\over{\|\langle\dot{\mu}_{k,t}, \dot{\mu}_{k',t}\rangle\|}} + {{\langle\ddot{\mu}_{k,t}, \ddot{\mu}_{k',t}\rangle}\over{\|\langle\ddot{\mu}_{k,t}, \ddot{\mu}_{k',t}\rangle\|}}\right).
    \label{eq:cosine_sim_function}
\end{equation}

Jointly with the proposed $\mathcal{L}_{traj}$, the affinity estimator finds $A_n$ that minimizes a loss function composed of following terms~\cite{bozic2021neural}: the graph local consistency loss $\mathcal{L}_{local}$, the graph time consistency loss $\mathcal{L}_{time}$, and the graph complexity loss $\mathcal{L}_{complex}$, which are
\begin{equation}
    \mathcal{L}_{local} = {1\over TK^2}\sum_t\sum_k\alpha_{k,t}\sum_{k'}a_{kk'}l_{t,kk'}
    \label{eq:local_consistency_loss}
\end{equation}
\begin{equation}
     \mathcal{L}_{time} = {1\over TK^2}\sum_t\sum_k\alpha_{k,t}\sum_{k'}a_{kk'}(l_{t,kk'}-\bar{l}_{t,kk'})
    \label{eq:time_consistency_loss}
\end{equation}
\begin{equation}
     \mathcal{L}_{complex} = \sum_n\sum_{n'\neq n}\|A_n\odot A_{n'}\|_F.
    \label{eq:complex_loss}
\end{equation}
$\mathcal{L}_{local}$ and $\mathcal{L}_{time}$ are designed to enforce the proximity and the temporal invariance of the neighbors in Euclidean space, while $\mathcal{L}_{complex}$ helps the neighbors of each keypoint to be different by minimizing the Frobenius norm of the Hadamard product of $A_n$ and $A_{n'}$.

\paragraph{Skeleton extraction}
The forward kinematics described in Eq.~(\ref{eq:fk}) assumes a tree structure, which starts from the root and progressively applies relative rotations on the joints of bones.
After the affinity matrix is found, we choose the minimal number of edges with high affinity values that create a single connected component of keypoints.
Then we find the root from the connectivity information, choosing the keypoint that has the shortest distance to all the other keypoints. 
Once the root is defined, we can traverse the tree and find links of parent-child relationship which can apply the forward kinematics of skeletal motion. The detailed algorithm to build the parent-child graph from the global affinity matrix is described in Sec. A.3 of the supplementary.
\subsection{Dynamics module}
Using the extracted skeletal topology, the dynamics module embeds the motion into the distribution in a latent space via standard encoder structure of a variational recurrent neural network (VRNN) with Eq.~(\ref{eq:loss_dynamics}).
While the conventional VRNN learns the latent variable $z_t$ that directly reconstructs the keypoints $\hat{x}_t$, our method encodes the local rotations of forward kinematics based on the skeletal topology.

Our decoder is implemented with the global pose decoder $\varphi_g$, and the rotation decoder $\varphi_r$.
The global pose decoder $\varphi_g$ decodes the translation of the root node $\hat{\mu}_{root,t}$ and the intensities $\hat{\alpha}_{k,t}$
\begin{equation}
     \hat{\mu}_{root,t}, \{~\hat{\alpha}_{k,t}\}_K = \varphi_g(z_t, h_t),
    \label{eq:global_pose_decoder}
\end{equation}
while the rotation decoder $\varphi_r$ extracts the relative rotations
\begin{equation}
     \{\tilde{R}_{k, t}\}_K = \varphi_r(z_t, h_t).
    \label{eq:rotation_decoder}
\end{equation}
The positions of keypoints $\hat{\mu}_{k,t}$ are recovered from the forward kinematics process of Eq.~\eqref{eq:fk}, and full reconstructions $\hat{x}_t = \{(\hat{\mu}_{k,t}, \hat{\alpha}_{k,t})\}_K$ are trained to minimize Eq.~\eqref{eq:vrnn_decoder}.

We additionally propose a randomized method to mitigate the difficulty in defining the canonical relative rotation. 
Training with the forward kinematics requires the canonical pose of the given skeleton, which is unknown during our unsupervised setting.
The canonical pose is also referred to as A-pose or T-pose, and is known a priori to define consistent relative rotations from the observation of $x_t$.
Specifically, the canonical pose defines $d_k$ in Eq.~(\ref{eq:offset}) and has to be shared for all episodes of data with the same topology of skeleton.

We suggest to randomly fix the orientation of each offset ${\bar{d}_k={d_k\over\|d_k\|}}\in\mathbb{R}^3$ at the beginning of the training step.
Then, the complete offset $d_k$ is simply scaled from $\bar{d}_k$ by the length of a bone detected in the first frame,
\begin{equation}
    \begin{gathered}
        d_k = \bar{d}_k\|\mu_{k,1} - \mu_{ parent(k),1}\|_2.
    \end{gathered}\label{eq:offset_scaling}
\end{equation}
The proposed randomized orientation is crucial to stabilize the training of the motion dynamics.
We validate the estimated rotation in the motion retargeting task.

The dynamics module of Neural Marionette successfully embeds the motion semantics with kinematics chain used in animation or robot control, while it defines the loss in the explicit physical space.
We demonstrate that our dynamics module effectively captures the distribution of motions that can generate plausible motion for various tasks.

\section{5\quad Experiments}

Our approach extracts the skeleton of unknown topology, and we show the generalization with a wide variety of targets: \textbf{D-FAUST}~\cite{bogo2017dynamic} and \textbf{AIST++}~\cite{li2021learn} for humans, \textbf{HanCo}~\cite{zimmermann2019freihand,zimmermann2021contrastive} for human hands, and \textbf{Animals}~\cite{li20214dcomplete} for various animals.
We also generated a sequence of a dynamic motion of a robot arm, \textbf{Panda} with the physics-based robot simulator~\cite{rohmer2013v}.
We randomly assigned episodes in dataset into the train and test split such that the ratio of total number of frames is roughly 9:1.
For quantitative evaluations, we extract a number of randomly cropped sequences for each episode in the test set, and average the results to obtain the final score of the corresponding episode.

\subsection{Skeletons}

\begin{figure}[t]
\centering
\includegraphics[width=0.45\textwidth]{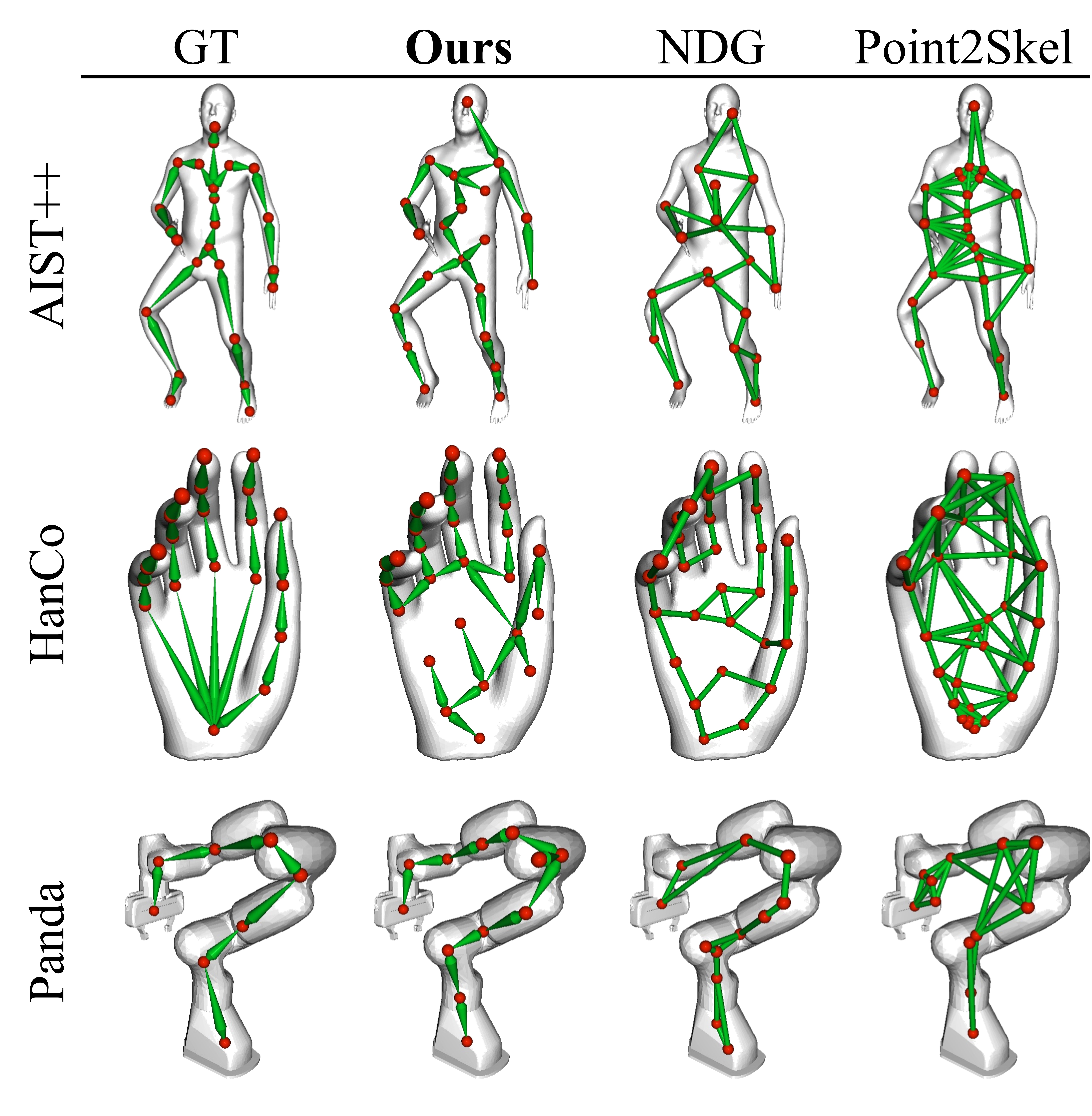} 
\caption{Qualitative comparison of extracted skeletons. Only the nodes whose intensities are above 0.2 are visualized for our model. Note that every edge has a parent-child relationship in ground truth and our model, while skeletons from NDG and Point2Skeleton do not have any hierarchy.}
\label{fig:skeletons}
\end{figure}

We first examine the discovered skeleton topology with \textbf{AIST++}, \textbf{HanCo}, and \textbf{Panda} dataset which contain the ground truth skeleton models. 
We compare the quality against two recent works that find skeletons from 3D observations in a fully unsupervised manner: Point2Skeleton ~\cite{lin2021point2skeleton} and Neural Deformation Graph (NDG)~\cite{bozic2021neural}. 
As Point2Skeleton is a model for extracting topology in a single frame of point cloud, we additionally augment the architecture with the spatio-temporal module of CaSPR~\cite{rempe2020caspr} for fair comparison.
With the temporal extension, both Point2Skeleton and our model can be optimized for the entire dataset to extract consistent topology.
On the other hand, NDG can only be optimized for a single episode.
We also would like to note that NDG uses a sequence of signed distance function grids and therefore observes richer information than point cloud. 

The recovered skeletons are compared against the ground truth skeleton in Figure~\ref{fig:skeletons}.
While the unsupervised skeleton might not exactly coincide with the ground truth, we can see that our joints are nicely spread within the volume and the links align with rigid parts of fingers or limbs.
The number of nodes used for our model and baselines are 24 for \textbf{AIST++}, 28 for  \textbf{HanCo}, and 12 for \textbf{Panda} dataset.
Note that Neural Marionette finds the minimal skeletal graph that connects high-intensity nodes to best represent the motion.
This implies that it can readily be applied to sequences with an unknown skeleton as long as the initial number of nodes is sufficient.
The flexibility is an essential advantage of Neural Marionette to represent unknown motion, whereas other approaches are restricted to the fixed number of nodes.
Our skeleton is deduced solely from the observation without any prior information, and therefore can be applied to various dynamic entities including humans, hands, animals, and robots as depicted in Figure~\ref{fig:overview}.

\begin{table}[t]
\centering
\small
    \begin{tabular}{c|ccc}
         \toprule
        Models          & {AIST++} & {HanCo}  & {Panda}  \tabularnewline
        \midrule
        \textbf{Ours} & \textbf{0.804} (0.201)&  \textbf{0.944} (0.0946)  &  \textbf{0.954} (0.0952)   \\
        P2S    & 0.755 (0.150)  &  0.847 (0.161)    &  0.937 (0.133)\\
        \bottomrule
    \end{tabular}
\caption{Semantic consistency score from \textbf{AIST++}, \textbf{HanCo}, and \textbf{Panda} dataset. Values inside parenthesis denote standard deviation for each keypoint.}
\label{table:semantic_score}
\end{table}
\begin{table}[t]
\centering
\small
    \begin{tabular}{c|ccc}
         \toprule
        Models          & {AIST++} & {HanCo} & {Panda}  \tabularnewline
        \midrule
        \textbf{Ours}   & 3.42 (0.924)             & \textbf{1.97} (0.0932)  &\textbf{1.71} (0.458)\\
            GT          & \textbf{3.11} (1.03)    & 2.12 (0.0971)                       &1.77 (0.437)\\
        \bottomrule
    \end{tabular}
\caption{Chamfer distance($\times10^4$) between ground truth (GT) and reconstructed point sets that are sampled from voxel stream on 4D tracking. Values inside the parenthesis denote the 95\%-confidence interval.}
\label{table:tracking_result}
\end{table}

To measure the quality of the recovered skeleton, we introduce semantic consistency score (SC-score).
If the skeleton correctly reflects the deformation, the relative positions of nodes should be consistent with respect to the ground truth semantic labels even if the detailed topologies are different. 
SC-score simply indicates how consistent the closest nodes in the ground truth are for each joint in the recovered skeleton, which is represented as
\begin{equation}
    \text{SC}={1\over J}\sum_j\max_k p_j(k),
    \label{eq:sc-score}
\end{equation}
where $J$ is the number of nodes in the ground truth, and $p_j(k)$ is the observed probability of $k$-th keypoint being the nearest neighbor of $j$-th ground truth joint.
Table~\ref{table:semantic_score} shows that our skeleton outperforms Point2Skeleton (P2S) in every dataset and therefore reflects the correct topology and relative motion.
We exclude NDG for the calculation since NDG needs to be separately optimized for every episode.

\begin{figure}[t]
\centering
\includegraphics[width=0.48\textwidth]{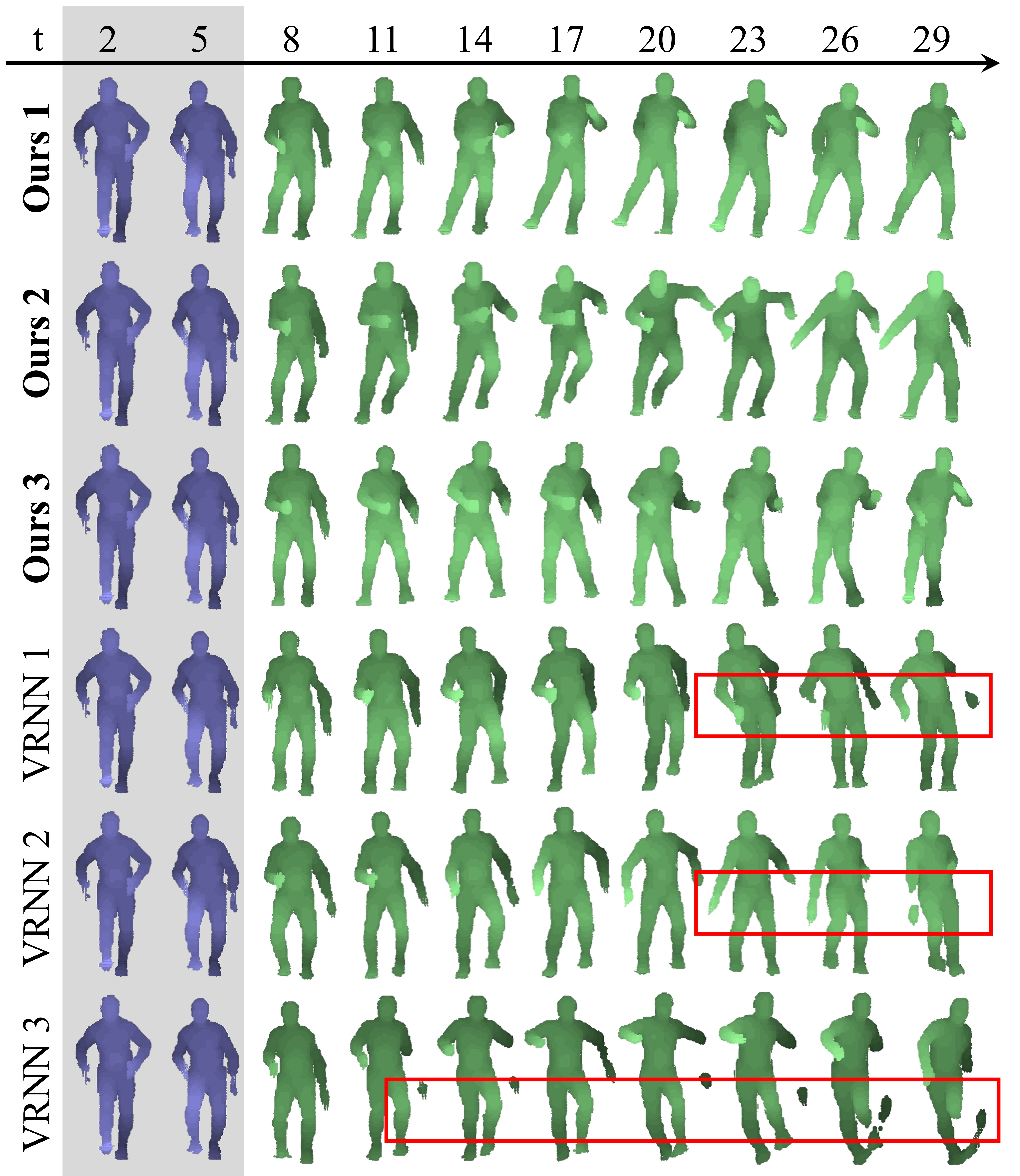} 
\caption{Result of motion generation on \textbf{AIST++} dataset with five conditioned frames. Stochastic latent variables from our model can generate multi-modal future frames, which are more plausible than the results from VRNN.}
\label{fig:motion_generation}
\end{figure}

In addition, we observed that the hand-labeled ground truth skeletons are not necessarily optimized to represent the motion.
For the baseline that is supervised with ground truth joints, we modified $\mathcal{L}_{vol}$ to minimize mean squared error between ground truth joints and the keypoints.
Although the same number of nodes are used, nodes from our model show results comparable to, or sometimes even outperform the ground truth in reconstructing the given 3D sequence (Table~\ref{table:tracking_result}). 
We can therefore conclude that our unsupervised skeleton can effectively capture the low-dimensional dynamics of volumetric sequence, which is further verified with various motion generation tasks.

\begin{figure}[t]
\centering
\includegraphics[width=0.48\textwidth]{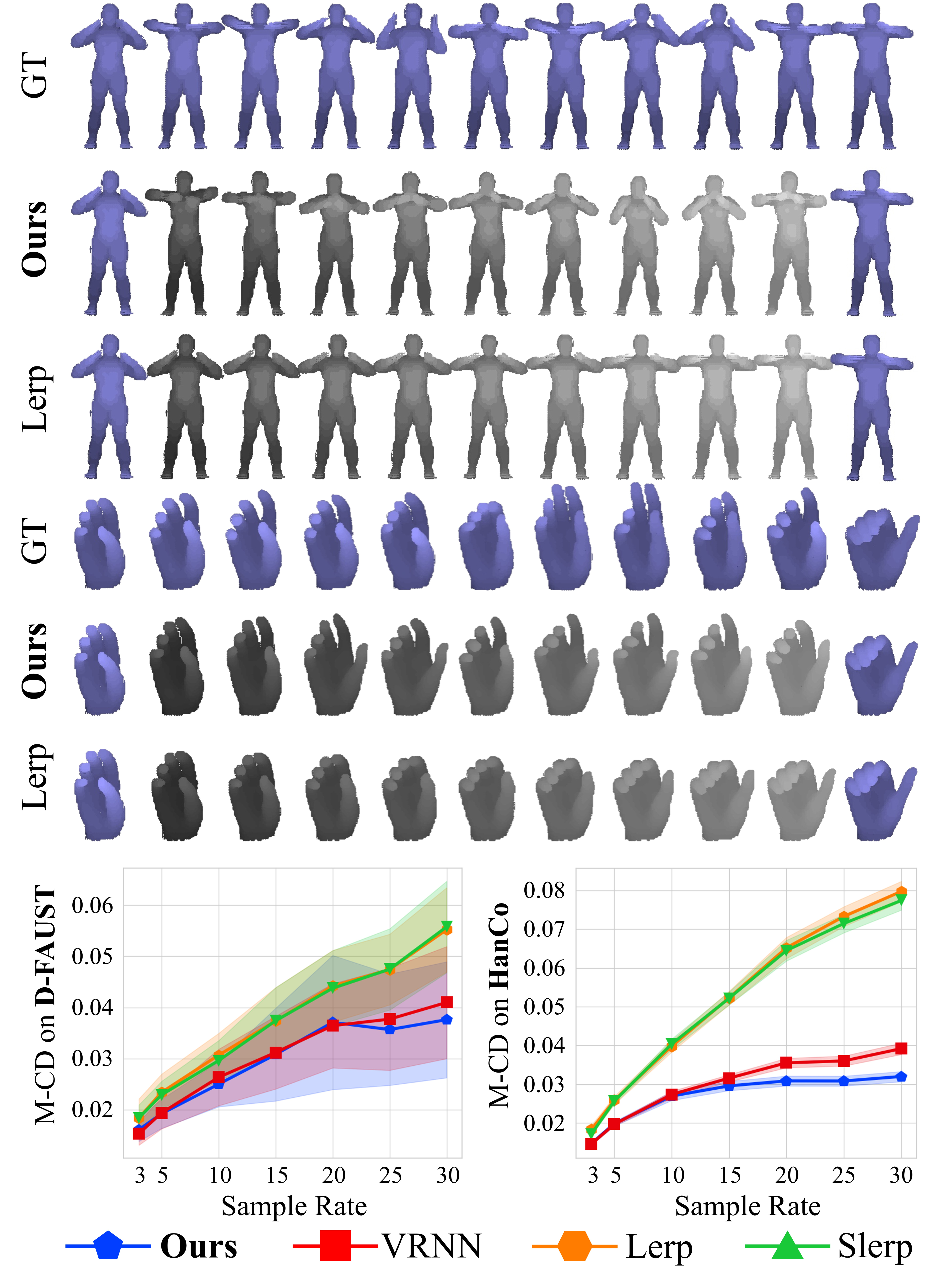} 
\caption{Result of motion interpolation on \textbf{D-FAUST} and \textbf{HanCo} dataset. Quantitative evaluations in Motion Chamfer Distance (M-CD) are plotted with 95\%-confidence interval.}
\label{fig:motion_interpolation}
\end{figure}

\begin{figure}[ht]
\centering
\includegraphics[width=0.48\textwidth]{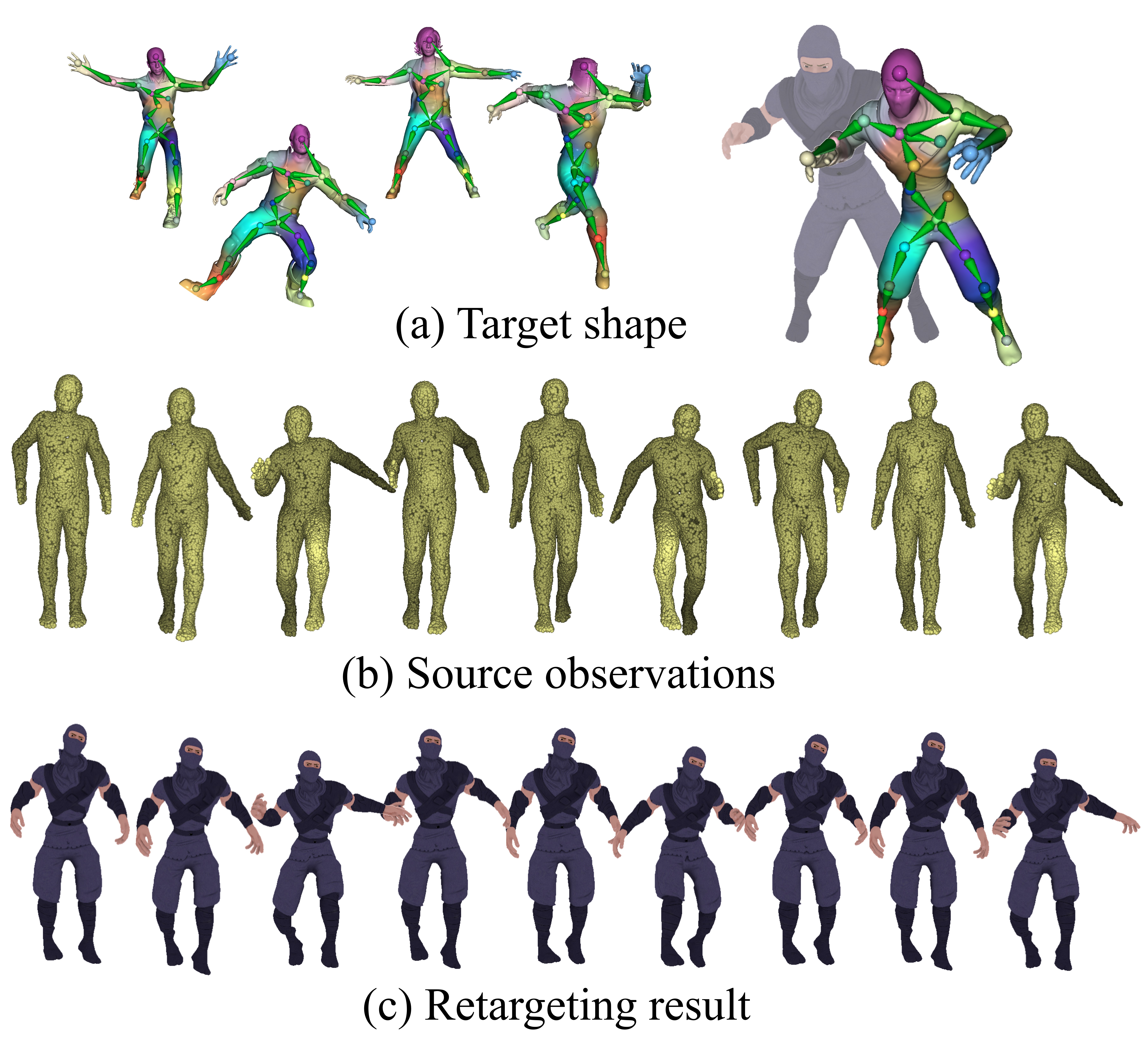} 
\caption{Overall process of motion retargeting. (a) Neural Marionette estimates the skeleton of an arbitrarily posed target shape, and (b) extracts skeletal motion from the source observations. (c) Then, the motion of the source object is retargeted into the target object.}
\label{fig:motion_retargeting}
\end{figure}
\subsection{Motion Generation}

The learned dynamics of Neural Marionette can generate a variety of plausible motion sequences.
Given $T_{cond}$ frames of observation, we test how the dynamics module can predict $T_{gen}$  frames of future sequences. 
We compare the quality of generated motion with the standard VRNN~\cite{chung2015recurrent,minderer2019unsupervised} using the same input skeleton.
Neural Marionette uses the tree structure and deduces the joint locations by learning the relative joint rotations, whereas VRNN directly regresses to the positions of the nodes.
The qualitative comparison with \textbf{AIST++} dataset is presented in Figure~\ref{fig:motion_generation}, where $T_{cond}$ is 5 and $T_{gen}$ is 25.
We find that our method creates much more natural and diverse motion compared to VRNN.
We argue that the chain of forward kinematics of the correct skeletal representation is simple yet crucial to correctly encoding the plausible motion sequence.
The individual regression of VRNN, on the other hand, results in inconsistent global positions and suffers from detached or prolonged parts.
More results are available in Sec. D.2 of the supplementary material.

\subsection{Motion Interpolation}

We can also interpolate the starting and ending poses of keyframes, given as $x_{t_s}$ and $x_{t_e}$, respectively.
For motion interpolation, we generate motion as previous section without additional optimization for the different task.
We sample latent variables $z_t$ from the posterior distribution $q_\phi$ of the starting frame $x_{t_s}$, and generate $N$ frames by sampling from prior distribution $p_\theta$.
We sample multiple trajectories of sequences, and select the one that ends with the pose closest to $x_{t_e}$. 
We can adapt a similar baseline that learns positional prior (VRNN).
We also added two non-generative baselines which either linearly interpolates the joints locations of the two poses (Lerp) or spherically interpolates the local joint rotations inferred from Neural Marionette (Slerp).

Figure~\ref{fig:motion_interpolation} (top) visually shows that our dynamics module more effectively interpolates the poses than baselines for \textbf{D-FAUST} and \textbf{HanCo} dataset.
Sequences generated with our method better follow the ground truth motion with interesting variations in between, whereas Lerp simply moves straight to the target pose.
The result with Slerp is similar to Lerp, and additional visualizations for interpolation in both datasets are contained in the supplementary material.

We suggest Motion Chamfer Distance (M-CD) using the voxel differences for quantitative comparison of reconstructed sequence, which is represented as 
\begin{equation}
    \begin{gathered}
        \text{M-CD}={1\over{T-1}}\sum_{t=2}^T \text{CD}(V^+_{t}, \hat{V}^+_t) + \text{CD}(V^-_{t}, \hat{V}^-_t)\\
    \end{gathered}\label{eq:MCD}
\end{equation}
where $V_t^+$ and $V_t^-$ refer to sampled point clouds from the positive  and negative voxels of $v_t-v_{t-1}$, and CD refers to the Chamfer Distance~\cite{fan2017point}.
When comparing the motion sequences, simply comparing the collocated occupancy can be biased toward the large overlapping static region, especially when the moving part is small with dense temporal sampling.
This is because that the binary grid does not contain any correspondence information. 
Instead, we find the difference volume better represents motion and use it to compare with the reconstructed sequence.

The plots in Figure~\ref{fig:motion_interpolation} quantitatively compare the interpolated motion against the ground truth.
We can clearly see that our dynamics module outperforms all of the baselines in all of the datasets with various topologies and motions.
Compared to the simple interpolation of Lerp and Slerp, the generative models of ours and VRNN performs significantly better. 
The result indicates that learning the motion context of joints is definitely crucial to generate plausible motion.
Also our encoding with forward kinematics performs superior to directly encoding positions of keypoints with VRNN.

\subsection{Motion Retargeting}

We also show that the motion extracted from Neural Marionette can be transferred to a different shape with the same skeleton topology as illustrated in Figure~\ref{fig:motion_retargeting}.
Neural Marionette encodes the relative 3D rotations of joints, enabling explicit control of motion for a given skeleton tree.
We extract the source motion with pretrained Neural Marionette from \textbf{AIST++} dataset and detect the skeleton in humanoids in Mixamo dataset.
We use standard linear blend skinning (LBS) and distance-based skin weights to deform the given shape, which is explained in detail in Sec. A.4 of the supplementary.
Note that the ground truth skeleton and the initial canonical pose of either datasets are not known, and the poses in sequences are arbitrary.
This is a highly challenging scenario compared to the standard rigging procedure, where a hand-crafted character is provided in a T-pose for adding bones and skins for further processing.

\begin{figure}[ht]
\centering
\includegraphics[width=0.48\textwidth]{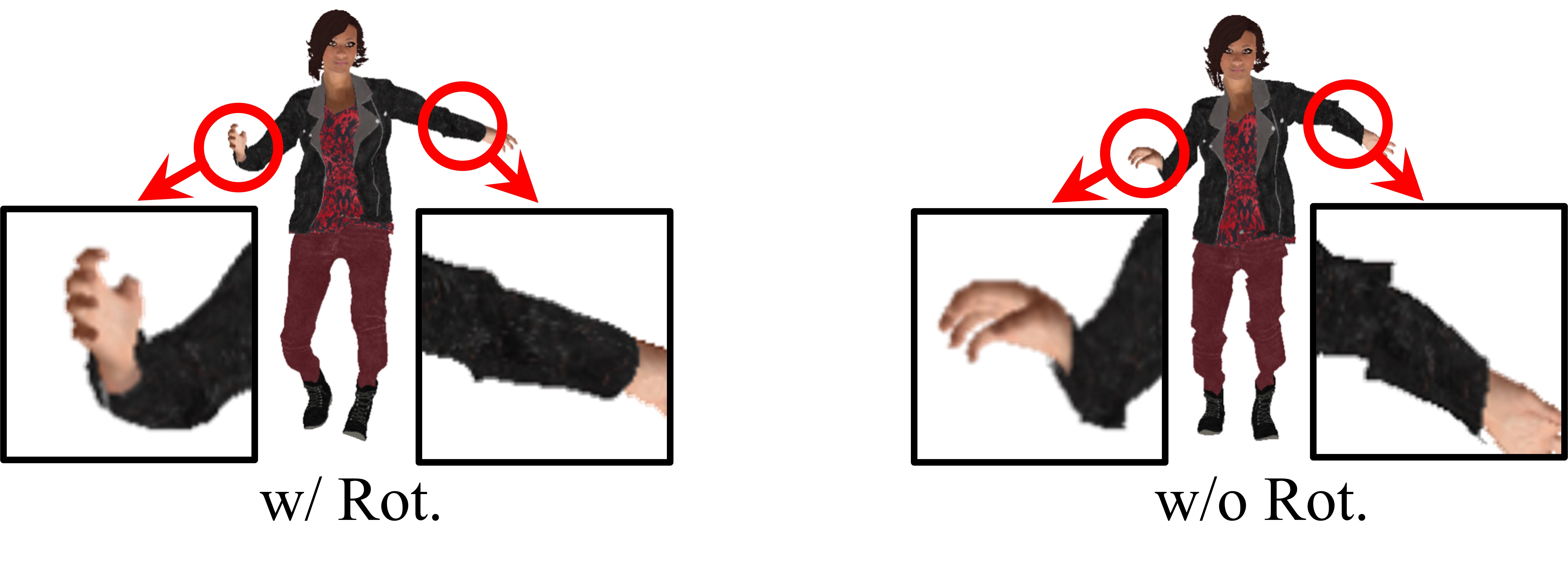} 
\caption{Comparison between motion retargeting results with and without rotations.}
\label{fig:retargeting_comparison}
\end{figure}

Neural Marionette deforms the mesh with the full relative transforms for forward kinematics creating natural motion. In contrast, the conventional motion representations with keypoints constrain only the locations of joints and therefore can create weird local rotations for joints.
The distortion induced from rotation mismatch is prominent when the motion is retargeted to textured characters as in Figure~\ref{fig:retargeting_comparison}.

\section{6\quad Conclusions}
In this work, we present Neural Marionette, an unsupervised approach that captures low-dimensional motion distribution of a diverse class of unknown targets without prior information.
Neural Marionette learns the motion dynamics and skeleton of a 3D model from a dynamic sequence and generates plausible motion from the learned latent distribution.
Our model is explicitly designed to apply the motion using forward kinematics equipped with a skeletal tree and corresponding per-joint rotation.
The direct relationship in the physical 3D space makes the representation highly interpretable.
The learned distribution is readily applicable for motion generation, interpolation, and retargeting without any fine-tuning for the specific tasks.
We believe that our work can be further expanded to a variety of tasks, from analyzing the movements of the unidentified target to generating plausible motions for many applications such as 3D character animation or autonomous agent control.
\bibliography{main}

\onecolumn
\LARGE
\begin{center}
    \textbf{Technical Appendix of\\
Neural Marionette: Unsupervised Learning of Motion Skeleton and 
\\Latent Dynamics from Volumetric Video}
\end{center}
\normalsize
\vspace{2em}

This document contains the implementation and training details with additional results that have been omitted due to the page limit in the main paper.
\section{A\quad Implementation Details}

Recall that Neural Marionette is composed of a \emph{skeleton module} and a \emph{dynamics module}.
The skeleton module is further decomposed into keypoint detection, affinity estimation, and skeleton extraction.
The keypoint detector of skeleton module and the dynamics module are deep neural networks trained with a dataset of volumetric sequences, whereas the affinity estimation and skeleton extraction are optimization and algorithmic steps respectively.
Sec. A.1. provides the detailed neural network architecture for the skeleton module, focusing on the keypoint detector, and the dynamics module.
Sec. A.2 shows the complete loss function used for the skeleton module.
Next, Sec. A.3. explains the detailed algorithm  to excavate the parent-child graph from the learned affinity matrix.
Sec. A.4 additionally describes how the skin weights are calculated for the given target mesh in the motion retargeting task.

\subsection{A.1\quad Detailed Architecture of Neural Marionette}


\begin{figure}[h]
\centering
\includegraphics[width=0.95\textwidth]{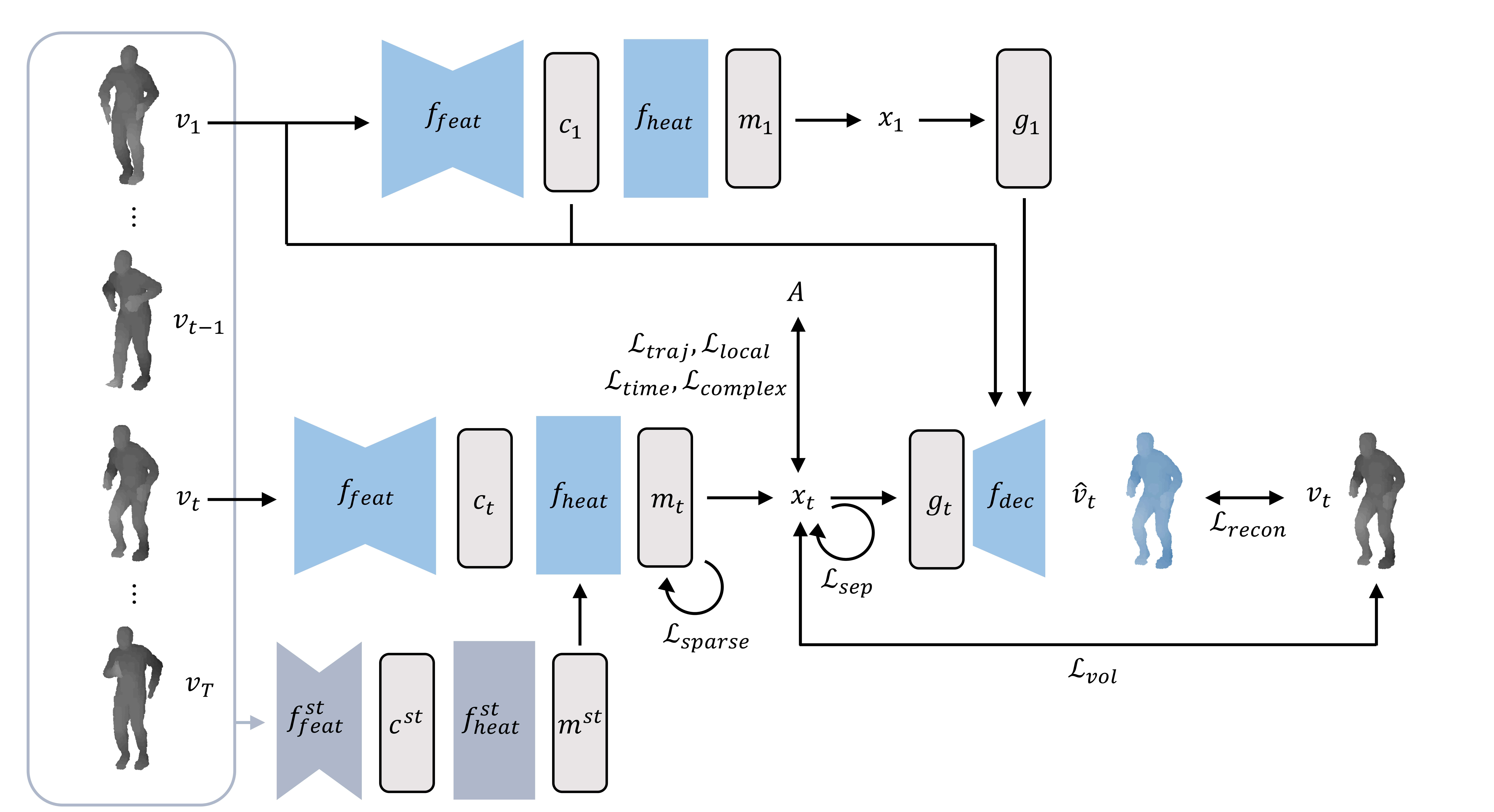} 
\caption{Architecture of skeleton module.}
\label{fig:skeleton_module_arch}
\end{figure}

\paragraph{Skeleton Module}
The suggested skeleton module consists of three building blocks: \emph{keypoint detector}, \emph{affinity estimation} and \emph{skeleton extraction}.
The keypoint detector finds the keypoints $x_t$ for a given voxel frame $v_t$ which serve as the candidate nodes of the skeleton.
After the affinity estimator learns the affinity matrix $A$, the skeleton extraction links the detected keypoints with directional edges that indicate the parent-child relationship.

Figure~\ref{fig:skeleton_module_arch} describes the network architecture and the training losses of the skeleton module.
Given a sequence of volumetric data $v_{1:T}$, the keypoint detector extracts the keypoints $x_t$ for individual frames.
The keypoint detector is basically composed of a set of 3D CNN-based autoencoder that embeds the input sequence into a low-dimensional keypoints and reconstructs the original sequence $\hat{v}_{1:T}$.

The encoder is composed of the feature extractor $f_{feat}: v_t \mapsto c_t$ and a heatmap extractor $f_{heat}: c_t \mapsto m_t$ from which the 3D coordinates and intensities of $x_t$ can be determined.
Note that $f_{feat}^{st}$, $f_{heat}^{st}$ in voxel encoder are the neural networks whose structures are totally identical to feature extractor $f_{feat}$ and heatmap extractor $f_{heat}$, respectively. 
The layers of the encoder extract spatio-temporal features from the temporal mean of the voxel sequence $v_{1:T}$, and contribute generating individual heatmaps $m_t$, which is abstracted in Eq. (6) of the paper.
The keypoints are converted to gaussian maps $g_t$ and decoded into the volumetric sequence $f_{dec}: g_t \mapsto \hat{v}_t$.
In practice, the decoder observes the first frame and reconstructs the difference with the first frame as mentioned in the main paper.

\begin{figure}[h]
\centering
\includegraphics[width=0.90\textwidth]{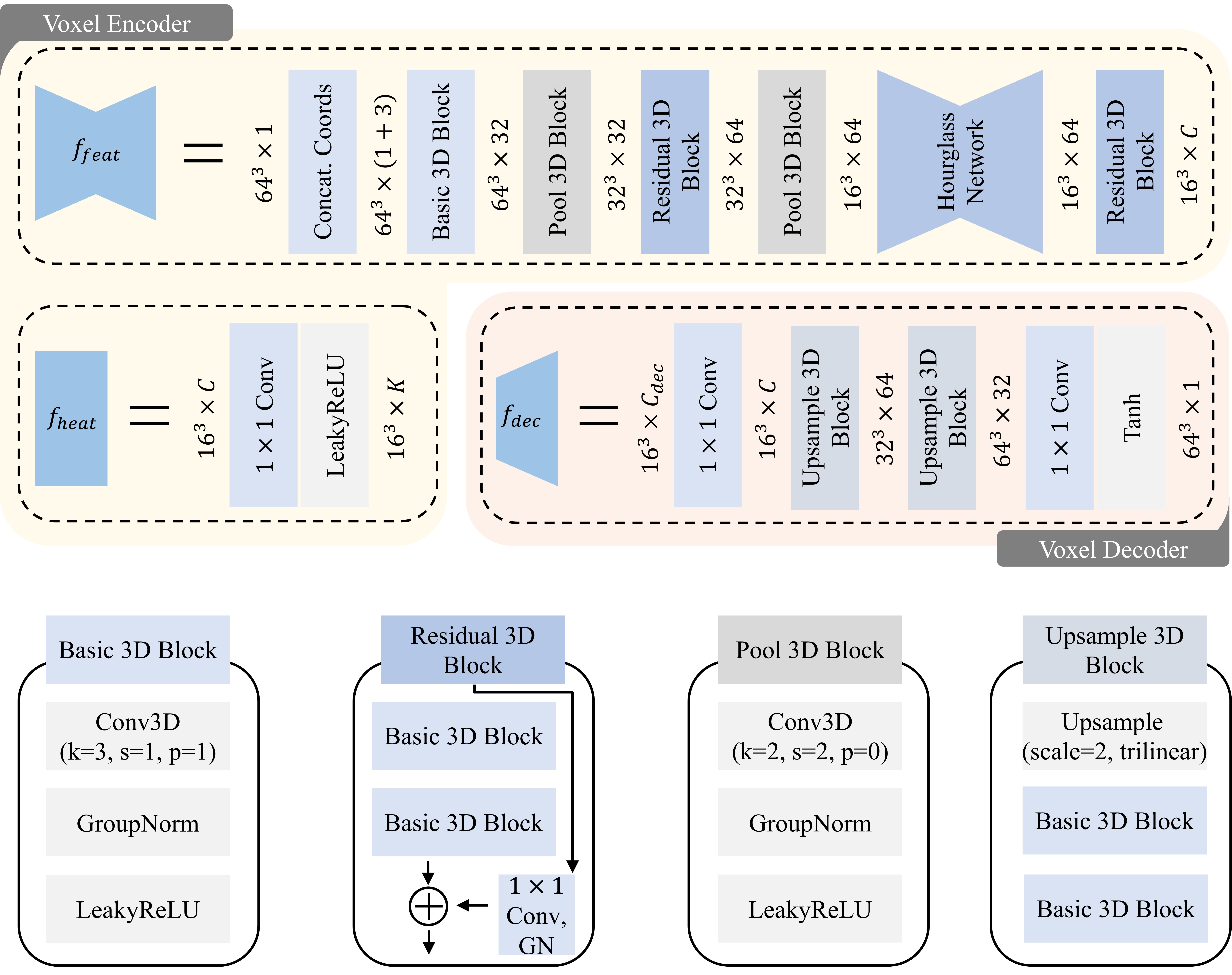} 
\caption{Detailed implementation of keypoint detector in our skeleton module. Note that input voxel grid is regarded as $64\times64\times64$, and $C_{dec}$ indicates concatenated dimension of $g_t, g_1, c_1$ and 3D coordinates of each voxel i.e., $C_{dec}=2\times K + C + 3$.}
\label{fig:skel_spec}
\end{figure}

We specify the implementation of feature extractor $f_{feat}$, heatmap extractor $f_{heat}$ and voxel decoder $f_{dec}$ in keypoint detector with Figure~\ref{fig:skel_spec}.
Similar to the state-of-the-art 2D keypoint detector~\cite{minderer2019unsupervised}, our keypoint detector basically imports the core idea of CoordConv~\cite{liu2018intriguing}, which is marked as \textbf{Concat. Coords} in the Figure~\ref{fig:skel_spec}. Also, we adopt an hourglass network for volumetric data from the previous work~\cite{xu2019predicting} and utilize the structure in the construction of $f_{feat}$, which is marked as \textbf{Hourglass Network} in the Figure~\ref{fig:skel_spec}. Voxel decoder $f_{dec}$ also adopts augmentation with the coordinate features, so that input is concatenated grid of gaussian maps $g_t, g_1$, feature grid from the first frame $c_1$, and the coordinates.

In contrast to keypoints, the global affinity matrix does not require any additional neural networks and is totally learned with regressions with a set of loss functions (Sec. 4, A.2).
Similarly, extraction of parent-child hierarchy is an algorithmic process (Sec. A.3) and therefore no neural network is required.

\begin{figure}[t]
\centering
\includegraphics[width=0.95\textwidth]{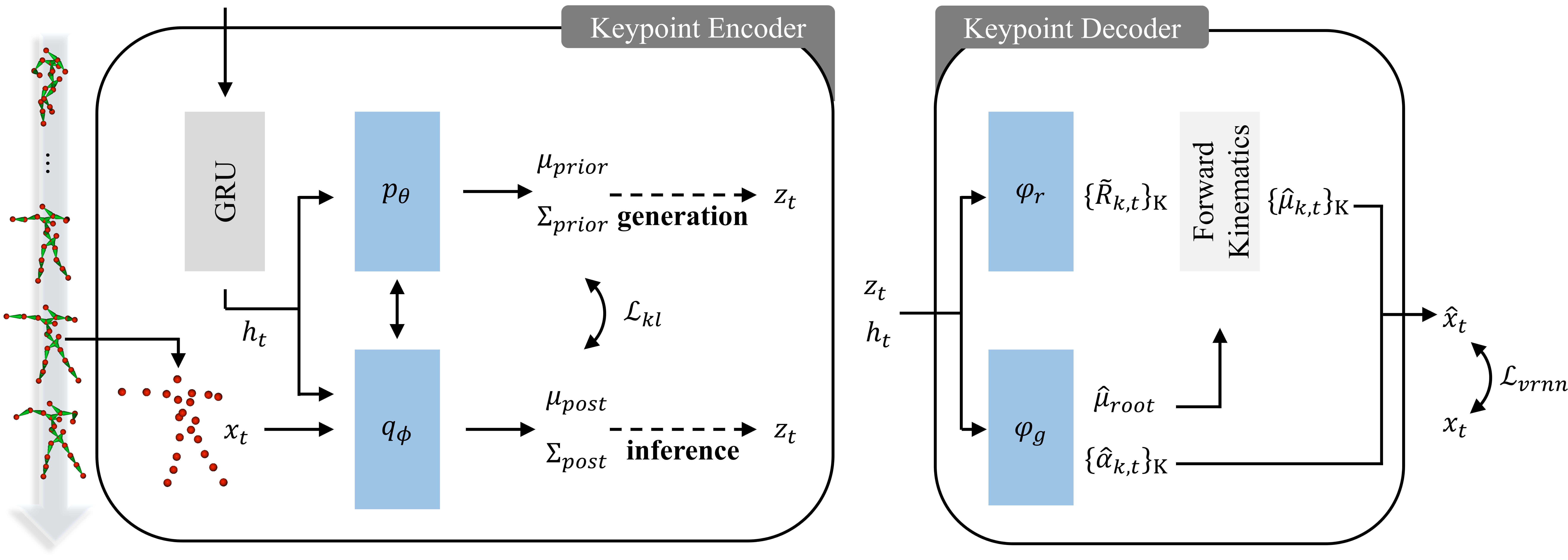} 
\caption{Architecture of dynamics module.}
\label{fig:dynamics_module_arch}
\end{figure}

\paragraph{Dynamics Module}
Dynamics module of Neural Marionette is based on the structure of variational recurrent neural network (VRNN)~\cite{chung2015recurrent} (Sec 3).
Overall architecture of our dynamics module is illustrated in Figure~\ref{fig:dynamics_module_arch}.

The nodes of the skeleton are encoded into the latent code in the encoder structure of dynamics module.
As encoder of our dynamics module follows the standard VRNN model, keypoints $x_t$ are encoded into the posterior distribution $q_\phi$ with the help of the hidden state $h_t$ and the stochastic latent variable $z_t$ is sampled from the $q_\phi$ during the inference.
At the same time, $h_t$ is soley encoded into prior distribution $p_\theta$ and $p_\theta$ tracks $q_\phi$ by minimizing KL-divergence loss $\mathcal{L}_{kl}$ explained in Eq. (4).
Note that $x_t$ here is a flattened form of $K$ keypoints so that it is a one-dimensional vector with the length of $K\times4$.

\begin{figure}[h]
\centering
\includegraphics[width=0.80\textwidth]{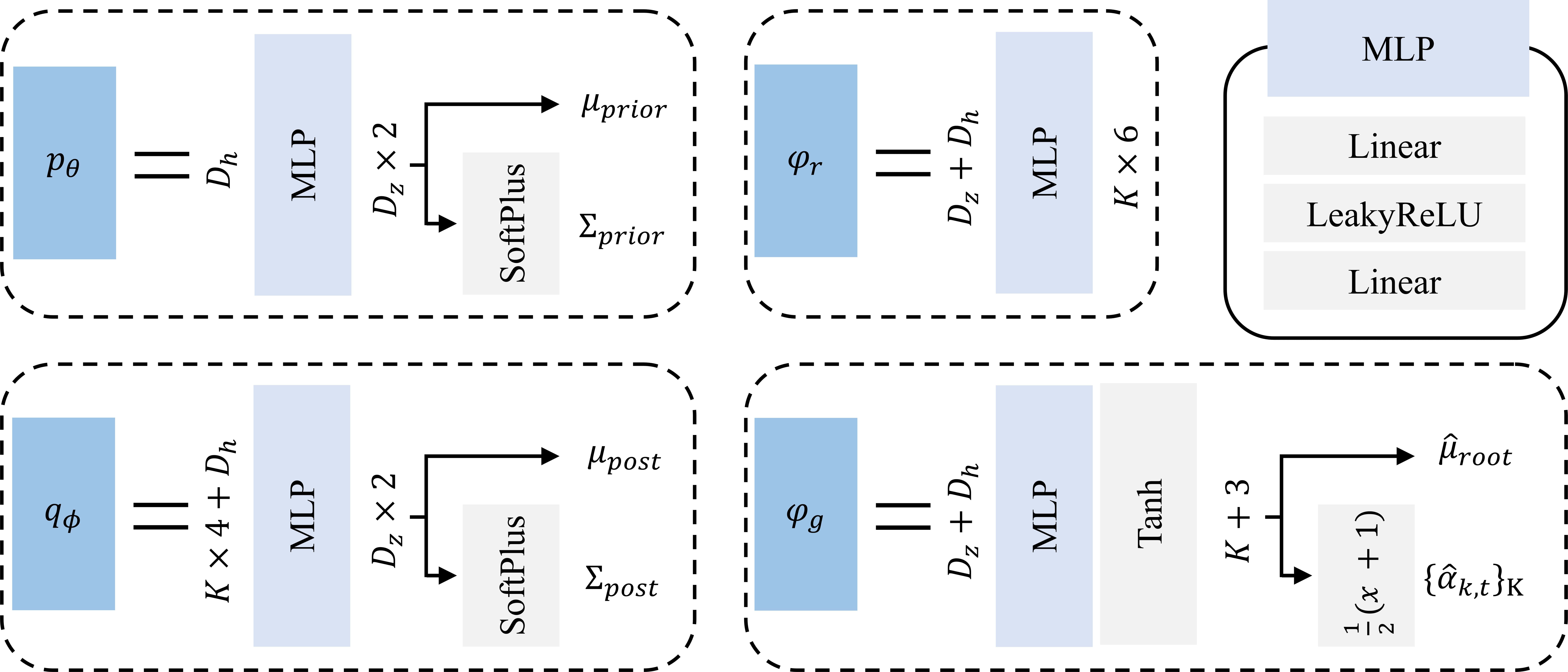} 
\caption{Detailed implementation of $p_\theta$, $q_\phi$, $\varphi_r$, and $\varphi_g$ in dynamics module. Note that $D_z$ and $D_h$ indicate dimensions of $z_t$ and $h_t$ respectively, and ${1\over2}(x+1)$ indicates affine transformation to match the scale of predicted intensity in a range of (0, 1).}
\label{fig:dyna_spec}
\end{figure}

Compared to the standard VRNN that directly reconstructs the input $x_t$, our decoder exploits the forward kinematics chain (Sec. 3).
From $z_t$, the decoder outputs  the relative rotations $\{\tilde{R}_{k,t}\}_K$  in the local coordinates with the 6D representations~\cite{zhou2019continuity}.
Detailed functionality of our keypoint decoder $\varphi$ is written in the Sec. 4 with Eq. (12) and (13).
We additionally provide implementation details on the posterior network $q_\phi$, prior network $p_\theta$, rotation decoder $\varphi_r$, and the global pose decoder $\varphi_g$ with Figure~\ref{fig:dyna_spec}.

\subsection{A.2\quad Loss Functions of Skeleton Module}
The nodes and edges of the skeleton are jointly learned from the keypoint detector and affinity estimator in skeleton module, and each submodule is optimized from the four different loss terms, respectively.

\paragraph{Keypoint detector}
As described in Sec. 4 of the paper, keypoint detector is an architecture that aims to find the candidate nodes of the complete skeleton.
To achieve detection on possible joints, we suggest to utilize the volume fitting loss $\mathcal{L}_{vol}$ of Eq.~\eqref{eq:volume_fitting_loss} to uniformly spread keypoints within the 3D object, along with the three loss functions that are widely used in training keypoint detectors: the reconstruction loss $\mathcal{L}_{recon}$, the sparsity loss $\mathcal{L}_{sparse}$, and the separation loss $\mathcal{L}_{sep}$, which are represented as Eq.~\eqref{eq:recon_loss} to Eq.~\eqref{eq:separation_loss}.
The arrows in Fig.~\ref{fig:skeleton_module_arch} indicate on which representation within the neural network the loss functions have effect on.

$\mathcal{L}_{recon}$ is a basic loss term on autoencoder structure that simply enforce reconstructed voxel $\hat{v}_t$ to track the ground truth voxel $v_t$. By minimizing  binary cross entropy (BCE) of $\mathcal{L}_{recon}$, grid features $c_t$ and 3D heatmaps $m_t$ in Eq. (5) and (6) are trained to encode not only the best position for a keypoint, but also the shape variation so that details of the voxel can be reconstructed in the decoder.
$\mathcal{L}_{sparse}$ and $\mathcal{L}_{sep}$ are adopted from previous works on the 2D keypoint detector~\cite{jakab2018unsupervised, minderer2019unsupervised}. While $\mathcal{L}_{sparse}$ improves sparsity of each heatmap $m_{k,t}$ so that prevent capturing excessive number of keypoints, $\mathcal{L}_{sep}$ enforces each of the possible keypoints to have different trajectories.

The previous loss functions with $\mathcal{L}_{recon}, \mathcal{L}_{sparse}$ and $\mathcal{L}_{sep}$ are not sufficient for joint estimation of keypoints and their affinities. 
Sec. C.1. includes the ablation study on $\mathcal{L}_{vol}$ and verifies that proposed $\mathcal{L}_{vol}$ serves a crucial role on training the skeleton module to find better keypoints.

\paragraph{Affinity estimation}
Affinity matrix $A=\{a_{ij}\}\in\mathbb{R}^{K\times K}$ is a weight matrix that represents connections between keypoints, i.e. \emph{bones} of skeleton as explained in Sec. 4. 
We suggest the graph trajectory loss $\mathcal{L}_{traj}$ in Eq.~\eqref{eq:graph_traj_loss} to build bones for the pair of nodes that have high correlation in the movements, which is represented with the cosine similarity between the velocities and accelerations.
Additional loss functions are adopted from the previous work on unsupervised rigging~\cite{bozic2021neural}:
the graph local consistency loss $\mathcal{L}_{local}$, the graph time consistency loss $\mathcal{L}_{time}$, and the graph complexity loss $\mathcal{L}_{complex}$. 
Note that integrated expression of $\mathcal{L}_{local}$ and $\mathcal{L}_{time}$ are named as $\mathcal{L}_{affinity}$, and $\mathcal{L}_{complex}$ is referred to as $\mathcal{L}_{sparsity}$ in the original literature, and we modified the names to avoid possible confusion with other loss terms.
Also, since our keypoint has an intensity component $\alpha_{k,t}$, we slightly modify the original loss terms to ignore keypoints that with low intensity as written in Eq.~\eqref{eq:local_consistency_loss} to Eq.~\eqref{eq:complex_loss}.

\paragraph{Total loss functions}
In summary, the total objective function $\mathcal{L}_{skel}$ to train the neural network layers of our skeleton module is the weighted sum of eight loss functions
\begin{equation}
\begin{gathered}
        \mathcal{L}_{skel} = 
        \lambda_{vol}\mathcal{L}_{vol} +
        \lambda_{recon}\mathcal{L}_{recon} +
        \lambda_{sparse}\mathcal{L}_{sparse}
        + \lambda_{sep}\mathcal{L}_{sep}\\~~~~~~~~~~~~~~~~~~~~~~~~~~~~~
        + \lambda_{traj}\mathcal{L}_{traj} +
        \lambda_{local}\mathcal{L}_{local}+
        \lambda_{time}\mathcal{L}_{time} +
        \lambda_{complex}\mathcal{L}_{complex}.
    \end{gathered}
    \label{eq:total_loss_skel}
\end{equation}
The weights are hyper-parameters that differ in each dataset, and we specify the values in the description on hyper-parameter settings in Sec. B.2. 
\subsection{A.3\quad Affinity matrix to Parent-child Graph}
In this chapter, we provide the detailed algorithm to extract the parent-child relationship solely from the learned affinity matrix $A$, as explained in Sec. 4 of the paper.
Once the global affinity matrix $A$ is learned from the dataset, we can extract the shared skeleton topology as described in Algorithm~\ref{alg:affinity2skel}.

\RestyleAlgo{ruled}
    \SetKwComment{Comment}{// }{}
    \SetKwInOut{Input}{Input}
    \SetKwInOut{Output}{Output}
    \SetKw{Break}{break}
    \DontPrintSemicolon
    \begin{algorithm}[hbt!]
    \caption{Affinity matrix to parent-child graph}\label{alg:affinity2skel}
        \Input {affinity matrix $A\in\mathbb{R}^{K\times K}$ with $N$ maximum neighbors}
        \Output {$parents$}
        \SetKwFunction{FDijk}{Dijkstra}
        \SetKwFunction{FWDijk}{WeightedDijkstra}
        \SetKwFunction{FBin}{BinarizeAffinity}
        \SetKwFunction{FRef}{RefineGraph}
        \SetKwFunction{FPos}{HigherRankNeighborSet}
        \SetKwFunction{FSame}{SameRankNeighborSet}
    
        \Comment*[l]{First, compose symmetric adjacency matrix}
        $A^{\mathrm{bin}}\gets$\FBin{$A$}\Comment*[r]{$A_{ij}\in\mathbb{R},~A^{\mathrm{bin}}_{ij}\in[0, 1]~\forall i,j$}
        $A^{\mathrm{dist}}\gets$ \FDijk{$A^{\mathrm{bin}}$}\Comment*[r]{$A^{\mathrm{dist}}_{ij}\in\{0, 1, \cdots,1\text{e}4\}~\forall i,j$ where 1e4 for no path}
        $r\gets\text{argmin}_i\sum_jA^{\mathrm{dist}}_{ij}$\Comment*[r]{root has the shortest path to the others}
        \;
        \Comment*[l]{Next, make sure all the nodes form a connected graph}
        \While{$\mathcal{G}=(\mathcal{V}\gets1:K,~\mathcal{E}\gets A^{\mathrm{bin}})~~\mathrm{is~~not}$ connected graph}{
            $k\gets\text{argmin}_{i\notin\mathcal{N}(root)}\sum_jA^{\mathrm{dist}}_{ij}$\Comment*[r]{identify another root from the other group}
            $A^{\mathrm{bin}}_{rk}, A^{\mathrm{bin}}_{kr}\gets 1$\Comment*[r]{connect $r$ and $k$}
            $A^{\mathrm{dist}}\gets$ \FDijk{$A^{\mathrm{bin}}$}\Comment*[r]{recompute topological distance}
            $r\gets\text{argmin}_i\sum_j A^{\mathrm{dist}}_{ij}$\Comment*[r]{update root}
            
        }
        \;
        \Comment*[l]{Refine $A^{\mathrm{dist}}$ based on affinity $A$, and extract $K$-lengthed}
        \Comment*[l]{integer list $rank$ which indicates priority on distance to root}
        $A^{\mathrm{dist}},~rank\gets$\FRef{$A$, $A^{\mathrm{bin}}, A^{\mathrm{dist}}$}\Comment*[r]{$a^{\mathrm{dist}}_{ij}\in\mathbb{R}~\forall i,j$}
        \;
        
        \Comment*[l]{Find list of parents}
        $parents\gets\{\}$\;
        \For{$i~\mathrm{in}~1:K$}{
            \eIf{$i==r$}{
                $parent[i]\gets r$\Comment*[r]{in case of root}
            }{
                $H_i\gets$\FPos{$i, rank, A^{\mathrm{bin}}$}\;
                $parent[i]\gets\text{argmin}_{j\in H_i}|rank[j]-rank[i]|$\Comment*[r]{normal case}
                $S_i\gets$\FSame{$i, rank, A^{\mathrm{bin}}$}\Comment*[r]{same rank neighbors}
                \If{$S_i\neq\O$}{
                    \For{$j\in S_i$}{
                        $H_j\gets$\FPos{$j, rank, A^{\mathrm{bin}}$}\;
                        $l\gets\text{argmax}_{k\in H_i\cap H_j}|rank[k]-rank[i]|$\;
                        \If{$A_{lj}>A_{li}$}{
                            $parent[i]\gets j$\;
                            \Break
                        }
                    }
                
                }

            }
        }
        \KwRet $parents$\;
    \end{algorithm}

The algorithm contains four main steps.
First, we initialize the connectivity and choose the root of the skeleton tree (\textbf{line 1 to 3}). 
Given a float-numbered affinity matrix $A$ which is asymmetric, we build a symmetric binary matrix $A^{\mathrm{bin}}$ that indicates the neighborhood relationship of the nodes.
Treating the acquired $A^{\mathrm{bin}}$ as adjacency, it represents a graph  $\mathcal{G}=(\mathcal{V}, \mathcal{E})$ with unweighted edges.
Then we calculate the pairwise distances of all nodes  $A^{\mathrm{dist}}$ using Dijkstra algorithm on the graph.
The root node is the node whose sum of distances to all other nodes is the minimum.

The second step ensures a connected skeleton that can capture the holistic motion dynamics (\textbf{line 5 to 10} in Alg.~\ref{alg:affinity2skel}).   
Since there is no explicit constraints on the values of $A$, it is possible for some of the nodes being isolated, which results in a skeleton with disconnected parts.

We incrementally add an edge to link disjoint components until we have a connected graph.
To find the edge to connect a pair of disconnected components, we identify the pseudo roots for each component which has the smallest sum of distances, and directly connect the roots. Then the distance matrix $A^{\mathrm{dist}}$ and the root $r$ are updated to reflect modified gra ph topology.

The third step is the \textbf{\textbf{RefineGraph}} function that refines the discovered choice of edges referencing the original affinity $A$ (\textbf{line 12} in Alg.~\ref{alg:affinity2skel}).
The initial graph $\mathcal{G}$ from the first step has been constructed upon a binarized matrix $A^{\mathrm{bin}}$ where all edges are treated equally.
While it is an efficient initialization, the nuances  of original affinities are lost with the hard decision.
We add the edge distances to the graph based on the original affinity $A$ and update $A^{\mathrm{dist}}$ by running the Dijkstra algorithm on the weighted graph.
The function also returns a list of properties named as $rank$, which contains the rank of each node that reflects how close the distance to the root is.
Basically it contains the order from 1 (root) to $K$ (the farthest from the root).

Finally we find the parent of each node in the graph to find the full chain of forward kinematics. 
Normally, the parent of $i$-th node should satisfy three rules. First the parent must be an element of the neighbor set $\mathcal{N}(i)$, and its rank needs to be higher than the rank of child, i.e., \emph{it needs to be closer to the root}.
At the same time, the parent node should be the node with the lowest rank among the higher-rank neighbor set $H(i)$ as the parent is the youngest among ancestors (\textbf{line 19 to 20} in Alg.~\ref{alg:affinity2skel}).
However, the neighbor $j$ with the same rank as the current node $i$ may exist, and we need to avoid the case in which the node has a parent-child relationship with a node that should have been a grandparent node.
In the case with the same-rank neighbors, we first find the highest ranked neighbor $l$ from the shared higher-rank neighbor set $H(i)\cap H(j)$, and if the affinity between $l$ and $j$ is higher than between $l$ and $i$, we define $j$-th node as the parent of $i$-th node (\textbf{line 21 to 31} in Alg.~\ref{alg:affinity2skel}).
\subsection{A.4\quad Calculation of Skin Weights}

For motion retargeting experiments in Sec. 5.4 of the paper, we need the skin weight of the mesh that adheres to the extracted skeleton.
We implemented a simple distance-based skin weight calculation as described in Algorithm~\ref{alg:skin_weight}.
After the skin weights are provided, the joint configuration deforms the shape of the target mesh using the linear blend skinning (LBS) method.

\RestyleAlgo{ruled}
    \SetKwComment{Comment}{// }{}
    \SetKwInOut{Input}{Input}
    \SetKwInOut{Output}{Output}
    \SetKw{Define}{Define}
    \SetKw{Continue}{continue}
    \DontPrintSemicolon
    \begin{algorithm}[hbt!]
    \caption{Calculation of skin weight}\label{alg:skin_weight}
        \Input {mesh vertices $p\in\mathbb{R}^{N_p\times3}$, keypoints $x=(\mu,\alpha)\in\mathbb{R}^{K\times 4}$, $parents$, $root$}
        \Output {skin weights $W\in\mathbb{R}^{N\times K}$}
        \Define{$\epsilon\gets\mathrm{THRESHOLD}$}\;
        \Comment*[l]{First, identify position of each bone}
        $invalids=\{i~~\mathbf{for}~~i~~\mathrm{in}~~1:K~~\mathbf{if}~~\alpha_i<\epsilon\}$\Comment*[r]{indices set of low-intensity keypoints}
        $b\gets\mathbf{0}_{K\times3}$\Comment*[r]{$b_i$ is position of each bone}
        \For{$i\in1:K$}{
            \eIf{$i == root$}{
                $b_i\gets\mu_i$\;
            }
            {
                \While{$parents[j]\in invalids$}{
                    $parents[j]\gets parents[parents[j]]$\;
                }
                $j\gets parents[i]$\;
                $b_i\gets{{\mu_i + \mu_j}\over 2} $\;
            }
        }
        \;
        \Comment*[l]{Find nearest bone of each vertex}
        \SetKwFunction{Fdist}{CalculateDistMatrix}
        $D\gets$\Fdist{$p, b$}\Comment*[r]{$D\in\mathbb{R}^{N\times K}$}
        \SetKwFunction{Fmask}{MaskInvalids}
        $D\gets$\Fmask{$D, invalids$}\Comment*[r]{make $k$-th column of $D$ infinity if $k\in invalids$}
        \SetKwFunction{Fgauss}{GaussianWeight}
        \For{$n\in1:N$}{
            $child\gets\text{argmin}_i D_{ni}$\;
            $parent\gets parents[child]$\;
            $W_{n}\gets$\Fgauss{$p_n, \mu_{child}, \mu_{parent}$}\Comment*[r]{calculate weights with gaussian kernel}
        }

        \KwRet $W$\;
    \end{algorithm}
\section{B\quad Training Details}
In this section, we provide training details including dataset processing (Sec. B.1), and hyperparameter settings (Sec. B.2) for each dataset.

\subsection{B.1\quad Dataset}
Neural Marionette is a generalizable algorithm as it does not exploit any of the category-specific prior knowledge.
We tested our model with a variety of targets: humans with \textbf{D-FAUST}~\cite{bogo2017dynamic} and \textbf{AIST++}~\cite{li2021learn}, human hands with \textbf{HanCo}~\cite{zimmermann2019freihand,zimmermann2021contrastive}, various animals with \textbf{Animals}~\cite{li20214dcomplete}, and robot arms with \textbf{Panda}~\cite{rohmer2013v}.
All of the listed datasets include time sequences of 3D mesh. We sampled 20k points from the mesh, which serve as the input point cloud for Neural Marionette.
Note that our method does not require additional pre-processing such as flood-fill algorithm.

\paragraph{Sequence crop}
Each of the dataset contains various length of episodes.
Due to the memory limit of GPU, we could not utilize all of the frames at once in training time.
Therefore we dynamically crop the sequence from the randomly chosen start frame of the original episode with the desired length $T$, and as a result, redundant sequences are rarely observable during training.
We believe this method improves the generalization capability as it allows to observe more diverse observations than when using a fixed set of segments that has been cropped in advance.
Hyperparameter $T$ is set as 10 and 20 during training of skeleton module and dynamics module respectively as explained in Sec. B.2.

\begin{figure}[h]
\centering
\includegraphics[width=0.85\textwidth]{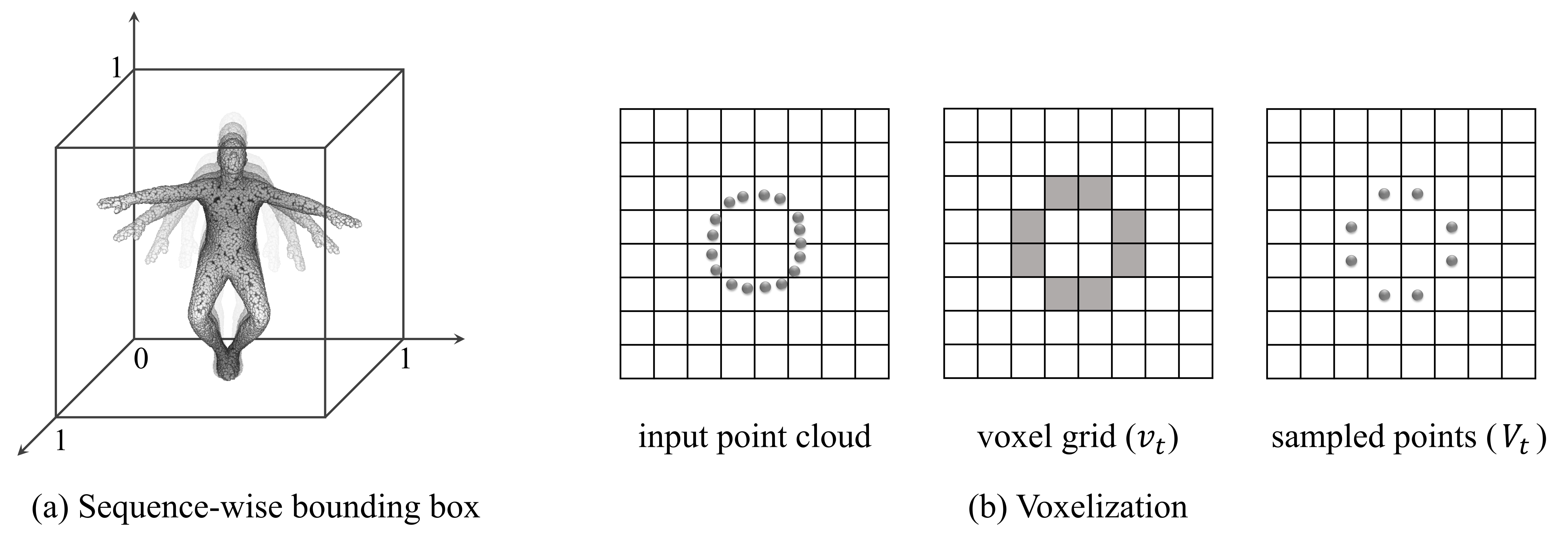} 
\caption{Visualization of voxelizing techniques. (a) Point cloud sequence is normalized in (0, 1) scale by employing the temporally shared bounding box for a given stream. (b) We additionally visualize the notations in the paper.}
\label{fig:voxelization}
\end{figure}

\paragraph{Voxelization}
By normalizing unstructured points in a fixed grid, voxelization enables understanding of the neighborhood context.
This is an important step for the skeleton module as it finds keypoints by observing changes in the neighboring voxels.
To maintain a fixed grid over the entire sequence of frames, we find the bounding box of points in all of the frames on a cropped sequence.
Then, the points of each frame are normalized along the detected bounding box, and are discretized into the voxel grid $v_t$ with the desired resolution as illustrated in Figure~\ref{fig:voxelization} (a).
Note that any points within the cell creates a occupied voxels, while the points sampled from the voxel is approximated to be at the center of the cell.
We utilize the sampled point cloud $V_t$ in the volume fitting loss $\mathcal{L}_{vol}$ of Eq. (8) to properly backpropagate the gradients, and also in the evaluation metrics for assessment on 4D tracking and motion interpolation of Eq.(16). 
The difference is visualized in Figure~\ref{fig:voxelization} (b).

\subsection{B.2\quad Hyperparameter Settings}
In this chapter, we provide additional information on hyperparameters for training Neural Marionette with each dataset.
General settings are included in Table~\ref{tab:general_hyper}, and tuned values for each dataset are written in the rest of the tables.
Note that training is conducted with GeForce RTX 3090 on Ubuntu 18.04 LTS system with the batch size given in Table~\ref{tab:general_hyper}.
Randomness in training is controlled with the pre-defined seed, which is also demonstrated in Table~\ref{tab:general_hyper}.
Some of the hyperparameters are manually tuned within a certain range: $K$ for (12, 32), $\sigma_g$ for (0.5, 2.0), $\lambda_{vol}$ for (0.1, 10.0), and all of the loss terms that are used in the affinity estimation for (1e-6, 10.0).

\begin{table}[htb!]
\caption{General hyperparameters}
    \centering
    \begin{small}
\begin{tabular}{lll}
\toprule
 Functionality & Notation & Value \\
\midrule
Grid size & $(G_x, G_y, G_z)$ & (64, 64, 64)\\
Optimizer &  & Adam \\
Random seed&  & 0\\
Learning rate decay [first, second] & & [0.25, 0.1] \\
Batch size for training  & $B_{skel}, B_{dyna}$ & 3, 16 \\
Length of the frames for training  & $T_{skel}, T_{dyna}$ & 10, 20\\
Maximum num. of neighbors & N & 2\\
Channels of feature grid $c_t$ &  C & 128 \\
Dimension of latent variable $z_t$ &  $\mathrm{D}_z$ & 128 \\
Dimension of hidden state $h_t$ & $\mathrm{D}_h$ & 512 \\
$\mathcal{L}_{sep}$ scale parameter & $\sigma_{s}$ & 1250 \\
Weight of $\mathcal{L}_{recon}$ &  $\lambda_{recon}$& 100.0\\
Weight of $\mathcal{L}_{sparse}$ &  $\lambda_{sparse}$& 5.0\\
Weight of $\mathcal{L}_{sep}$ &  $\lambda_{sep}$& 0.1\\
Weight of $\mathcal{L}_{time}$ &  $\lambda_{time}$& 1.0\\
Weight of $\mathcal{L}_{complex}$ &  $\lambda_{complex}$& 0.01\\
Weight of $\mathcal{L}_{vrnn}$ &  $\lambda_{vrnn}$& 1.0\\
Weight of $\mathcal{L}_{kl}$ &  $\lambda_{kl}$& 1.0\\

\bottomrule
\end{tabular}
\end{small}
    \label{tab:general_hyper}
\end{table}

\begin{multicols}{2}

\begin{table}[H]
\caption{Hyperparameters for \textbf{AIST++}}
    \centering

\begin{small}
\begin{tabular}{lll}
\toprule
 Functionality & Notation & Value \\
\midrule
Sample rate from ori. seq & & 2\\
Lr decay milestones (epoch) & & $[60, 140]$ \\
Num. of keypoints & K & 24\\
Sigma of gaussian & $\sigma_g$ & 1.5\\
Weight of $\mathcal{L}_{vol}$ &  $\lambda_{vol}$& 10.0\\
Weight of $\mathcal{L}_{traj}$ &  $\lambda_{traj}$& 1.0\\
Weight of $\mathcal{L}_{local}$ &  $\lambda_{local}$& 0.001\\
\bottomrule
\end{tabular}
\end{small}

    \label{tab:aist_hyper}
\end{table}

\begin{table}[H]
\caption{Hyperparameters for \textbf{D-FAUST}}
\centering
\begin{small}
\begin{tabular}{lll}
\toprule
 Functionality & Notation & Value \\
\midrule
Sample rate from ori. seq & & 5\\
Lr decay milestones (epoch) & & $[600, 1400]$ \\
Num. of keypoints & K & 24\\
Sigma of gaussian & $\sigma_g$ & 1.5\\
Weight of $\mathcal{L}_{vol}$ &  $\lambda_{vol}$& 10.0\\
Weight of $\mathcal{L}_{traj}$ &  $\lambda_{traj}$& 1.0\\
Weight of $\mathcal{L}_{local}$ &  $\lambda_{local}$& 0.001\\
\bottomrule
\end{tabular}
\end{small}
\label{tab:dfaust_hyper}
\end{table}

\begin{table}[H]
\caption{Hyperparameters for \textbf{HanCo}}
\centering
\begin{small}
\begin{tabular}{lll}
\toprule
 Functionality & Notation & Value \\
\midrule
Sample rate from ori. seq & & 1\\
Lr decay milestones (epoch) & & $[120, 170]$ \\
Num. of keypoints & K & 28\\
Sigma of gaussian & $\sigma_g$ & 1.0\\
Weight of $\mathcal{L}_{vol}$ &  $\lambda_{vol}$& 0.1\\
Weight of $\mathcal{L}_{traj}$ &  $\lambda_{traj}$& 1e-6\\
Weight of $\mathcal{L}_{local}$ &  $\lambda_{local}$& 1.0\\
\bottomrule
\end{tabular}
\end{small}
\label{tab:hanco_hyper}
\end{table}

\begin{table}[H]
\caption{Hyperparameters for \textbf{Panda}}
\centering
\begin{small}
\begin{tabular}{lll}
\toprule
 Functionality & Notation & Value \\
\midrule
Sample rate from ori. seq & & 1\\
Lr decay milestones (epoch) & & $[60, 140]$ \\
Num. of keypoints & K & 12\\
Sigma of gaussian & $\sigma_g$ & 1.5\\
Weight of $\mathcal{L}_{vol}$ &  $\lambda_{vol}$& 10.0\\
Weight of $\mathcal{L}_{traj}$ &  $\lambda_{traj}$& 0.001\\
Weight of $\mathcal{L}_{local}$ &  $\lambda_{local}$& 1.0\\
\bottomrule
\end{tabular}
\end{small}
\label{tab:panda_hyper}
\end{table}

\end{multicols}

\begin{table}[H]
\caption{Hyperparameters for \textbf{Animals}}
\centering
\begin{small}
\begin{tabular}{lll}
\toprule
 Functionality & Notation & Value \\
\midrule
Sample rate from ori. seq & & 1\\
Lr decay milestones (epoch) & & $[120, 170]$ \\
Num. of keypoints & K & 24\\
Sigma of gaussian & $\sigma_g$ & 2.0\\
Weight of $\mathcal{L}_{vol}$ &  $\lambda_{vol}$& 10.0\\
Weight of $\mathcal{L}_{traj}$ &  $\lambda_{traj}$& 1e-6\\
Weight of $\mathcal{L}_{local}$ &  $\lambda_{local}$& 0.001\\
\bottomrule
\end{tabular}
\end{small}
\label{tab:animals_hyper}
\end{table}

\section{C\quad Ablation studies on $\mathcal{L}_{vol}$ and $\mathcal{L}_{traj}$}

In this section, we verify the necessity of the volume fitting loss $\mathcal{L}_{vol}$ (Sec. C.1) and the graph trajectory loss $\mathcal{L}_{traj}$ (Sec. C.2).
We tested the result with the two challenging dataset for rigging: human models in \textbf{AIST++} and hand models in \textbf{HanCo}.

\subsection{C.1\quad Ablation on $\mathcal{L}_{vol}$}
As mentioned in Sec. A.2, $L_{vol}$ is a crucial loss term to bridge learning of keypoints and their affinities.
To demonstrate the effect of $L_{vol}$, we additionally train the model excluding $L_{vol}$, and compare it with our model that uses Eq.~\eqref{eq:total_loss_skel} as an objective.
Note that both models are trained with the equal number of epochs, which are 200 for \textbf{AIST++} and 150 for \textbf{HanCo} dataset.
The quantitative evaluations are conducted with the proposed SC-score of Eq. (15) and Chamfer distance to assess ability in 4D tracking, which are demonstrated in Table~\ref{table:semantic_score_L_vol} and Table~\ref{table:chamfer_L_vol}, respectively.

\begin{table}[ht]
\centering
    \begin{tabular}{c|cc}
         \toprule
        Models         & {AIST++} & {HanCo}  \tabularnewline
        \midrule
        \textbf{Ours}   & \textbf{0.804} (0.201)             & \textbf{0.944} (0.0946)\\
            w/o $\mathcal{L}_{vol}$          & 0.603 (0.265)    & 0.890 (0.150)\\
        \bottomrule
    \end{tabular}
\caption{Semantic score of keypoints from two models. Values inside parenthesis denote standard deviation for each keypoint.}
\label{table:semantic_score_L_vol}
\end{table}

\begin{table}[ht]
\centering
    \begin{tabular}{c|cc}
         \toprule
        Models          & {AIST++} & {HanCo}  \tabularnewline
        \midrule
        \textbf{Ours}   & \textbf{3.42} (0.924)             & \textbf{1.97} (0.0932)\\
         w/o $\mathcal{L}_{vol}$          & 16.7 (8.98)    & 2.30 (0.102)                       \\
        \bottomrule
    \end{tabular}
\caption{Chamfer distance($\times10^4$) between ground truth (GT) and reconstructed point sets that are sampled from voxel stream on 4D tracking. Values inside the parenthesis denote the 95\%-confidence interval.}
\label{table:chamfer_L_vol}
\end{table}
The results from the two metrics demonstrate that $\mathcal{L}_{vol}$ is essential for joint learning of keypoints and affinity matrix.
Especially for \textbf{AIST++} dataset which contains massive number of motions, training without $\mathcal{L}_{vol}$ significantly depress finding keypoints.
$\mathcal{L}_{vol}$ uniformly distributes the keypoints and provides a strong cues for joint locations of dynamic objects.

\subsection{C.2\quad Ablation on $\mathcal{L}_{traj}$}
\begin{figure}[ht]
\centering
\includegraphics[width=0.45\textwidth]{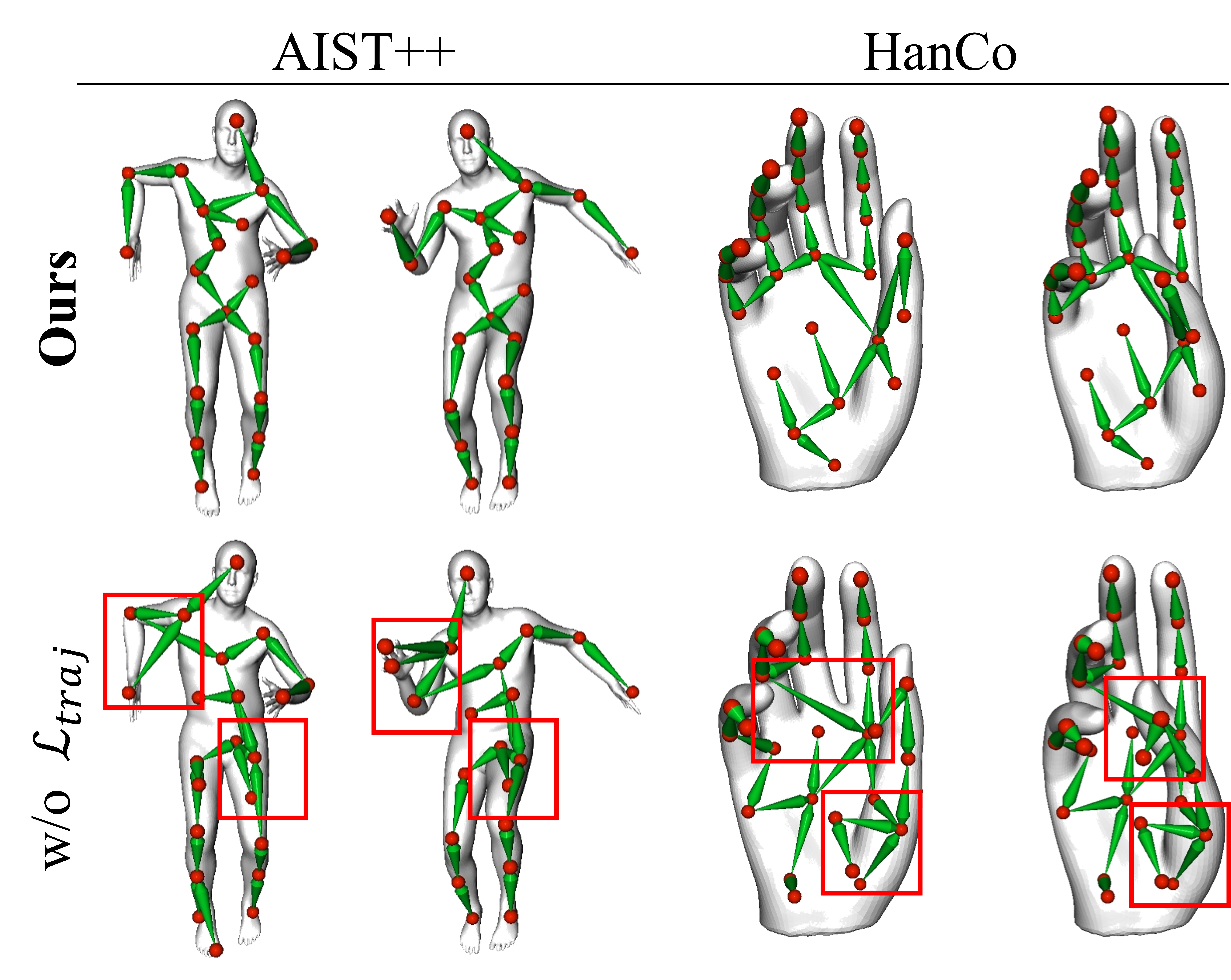} 
\caption{Qualitative comparison of extracted skeletons on \textbf{AIST++} and \textbf{HanCo} dataset.}
\label{fig:ablation_traj}
\end{figure}

We show that the proposed $\mathcal{L}_{traj}$ is essential to estimate affinity that correctly reflects the connections between a sparse set of joints. 
For affinity estimation, we suggest $\mathcal{L}_{traj}$ in addition to pre-existing losses $\mathcal{L}_{local}, \mathcal{L}_{time}$ and $\mathcal{L}_{complex}$.
We test the effectiveness of $\mathcal{L}_{traj}$ by training the skeleton module with and without $\mathcal{L}_{traj}$ for equal number of ephochs.
The resulting skeletons are compared visually in Figure~\ref{fig:ablation_traj}.
Although the positions of keypoints from the both models are almost identical, The bones of skeletons with the complete loss function better capture the rigid  parts of the body compared to the model trained without $\mathcal{L}_{traj}$.
This implies that $\mathcal{L}_{traj}$ contributes on extracting natural skeletons with accurate bones of the rigid parts.

\onecolumn
\section{D\quad Additional Samples}
In this section, we provide additional results on the experiments.
\subsection{D.1\quad Skeletons}

\begin{figure}[hbt!]
\centering
\includegraphics[width=0.85\textwidth]{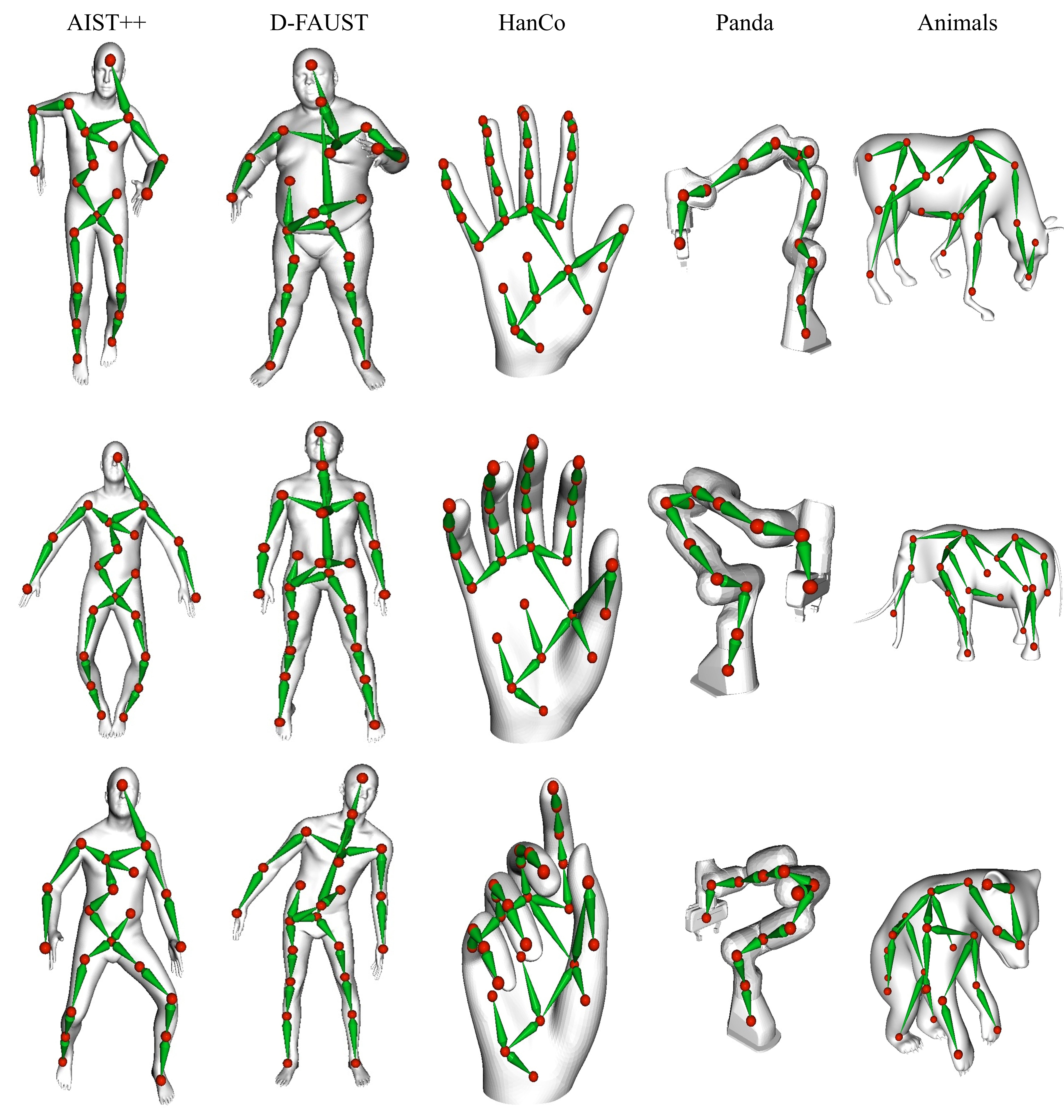} 
\caption{Skeletons extracted from Neural Marionette in all the dataset.}
\label{fig:additional_skels}
\end{figure}

\onecolumn
\subsection{D.2\quad Motion Generation}
\begin{figure}[hbt!]
\centering
\includegraphics[width=0.95\textwidth]{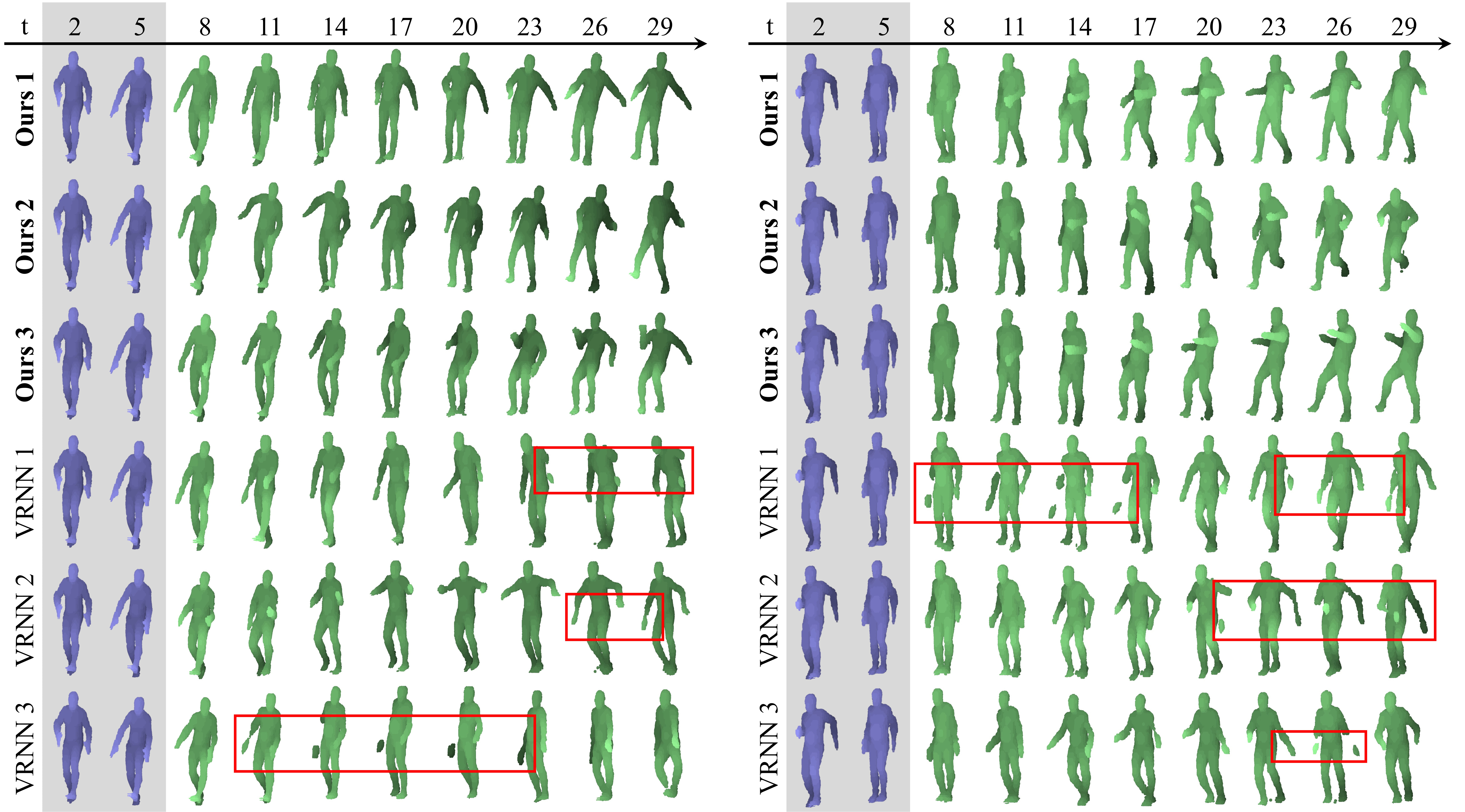} 
\caption{Additional samples for motion generation in \textbf{AIST++} dataset.}
\label{fig:additional_generation_aist1}
\end{figure}

\begin{figure}[hbt!]
\centering
\includegraphics[width=0.95\textwidth]{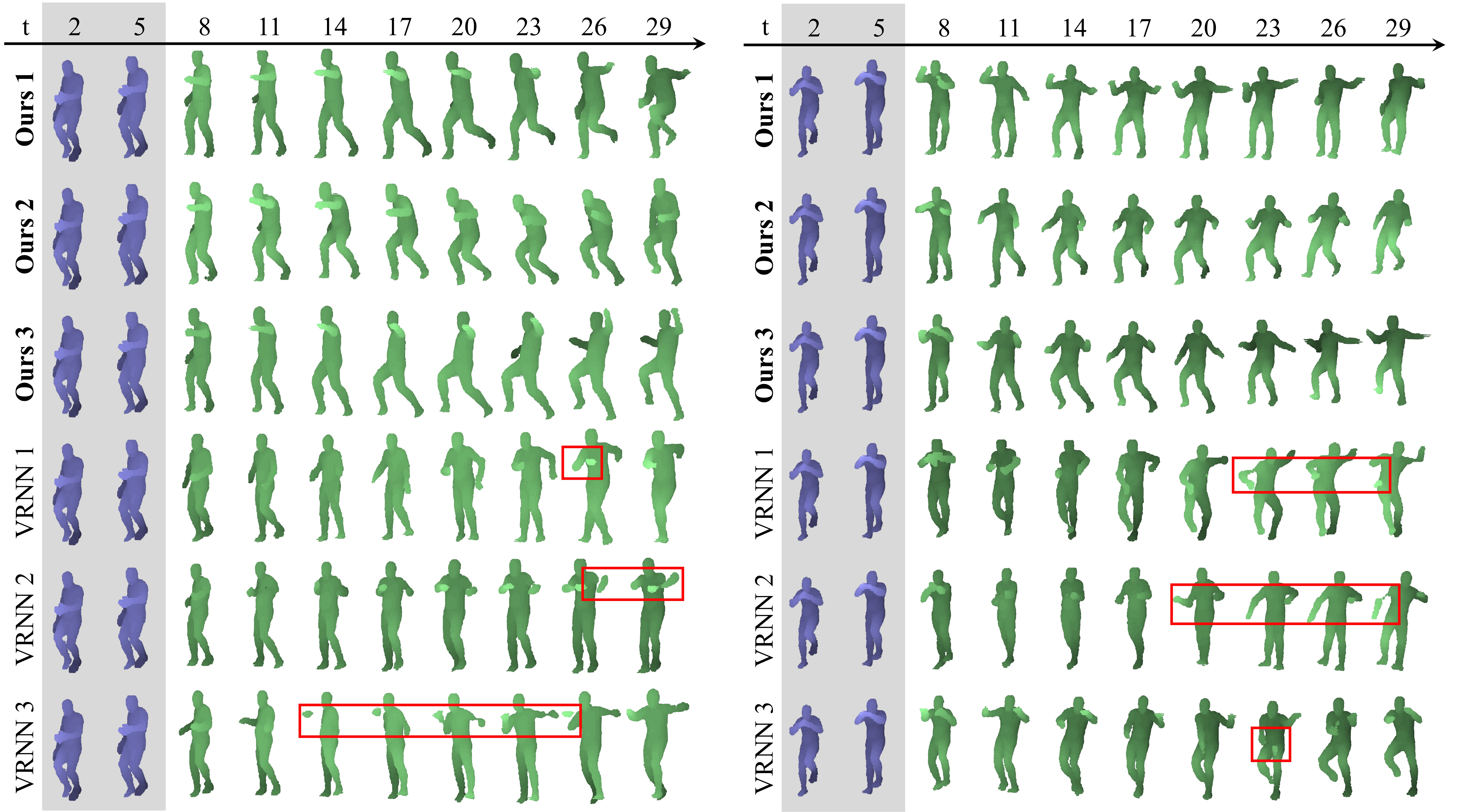} 
\caption{Additional samples for motion generation in \textbf{AIST++} dataset.}
\label{fig:additional_generation_aist2}
\end{figure}

\begin{figure}[hbt!]
\centering
\includegraphics[width=0.95\textwidth]{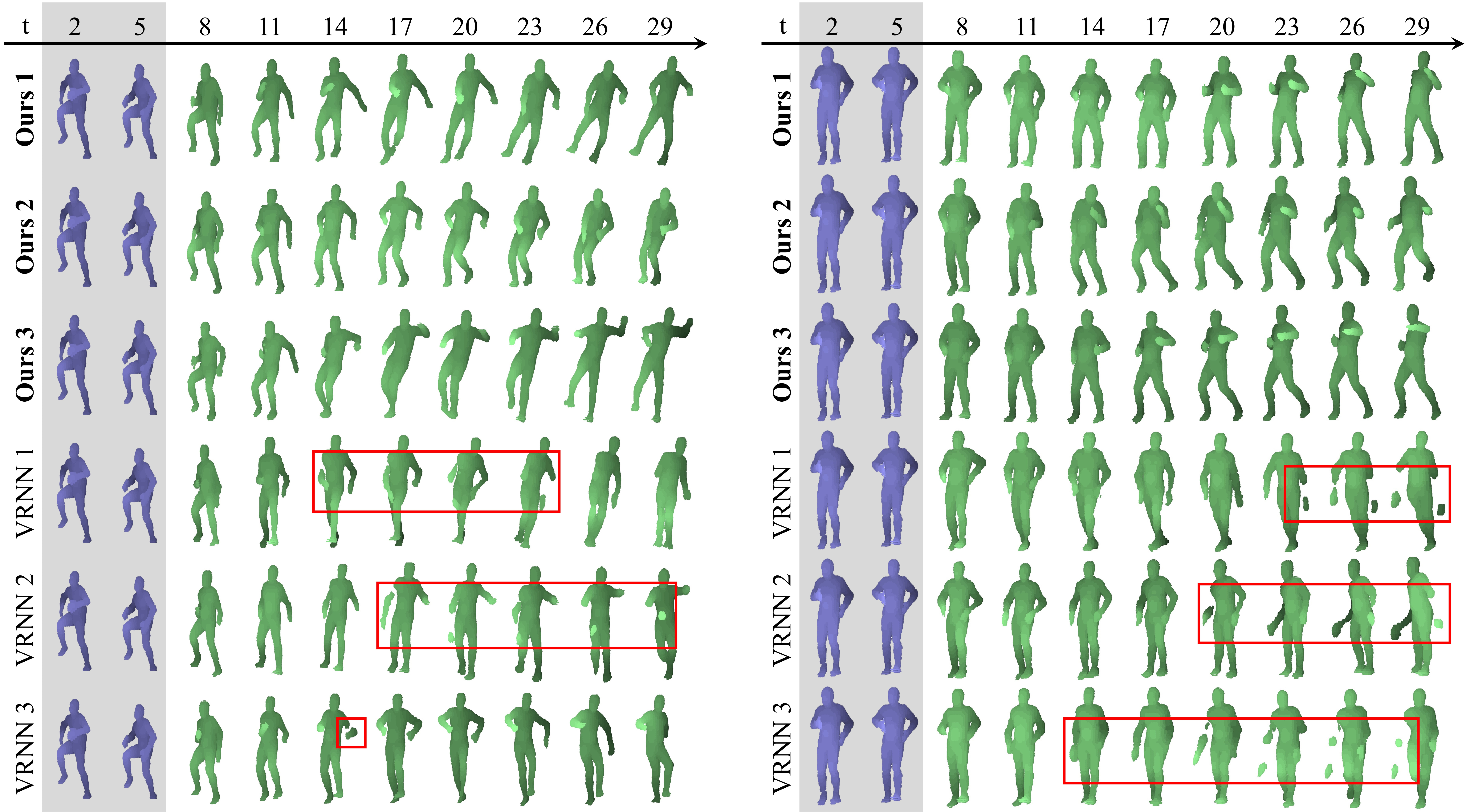} 
\caption{Additional samples for motion generation in \textbf{AIST++} dataset.}
\label{fig:additional_generation_aist3}
\end{figure}

\begin{figure}[hbt!]
\centering
\includegraphics[width=0.70\textwidth]{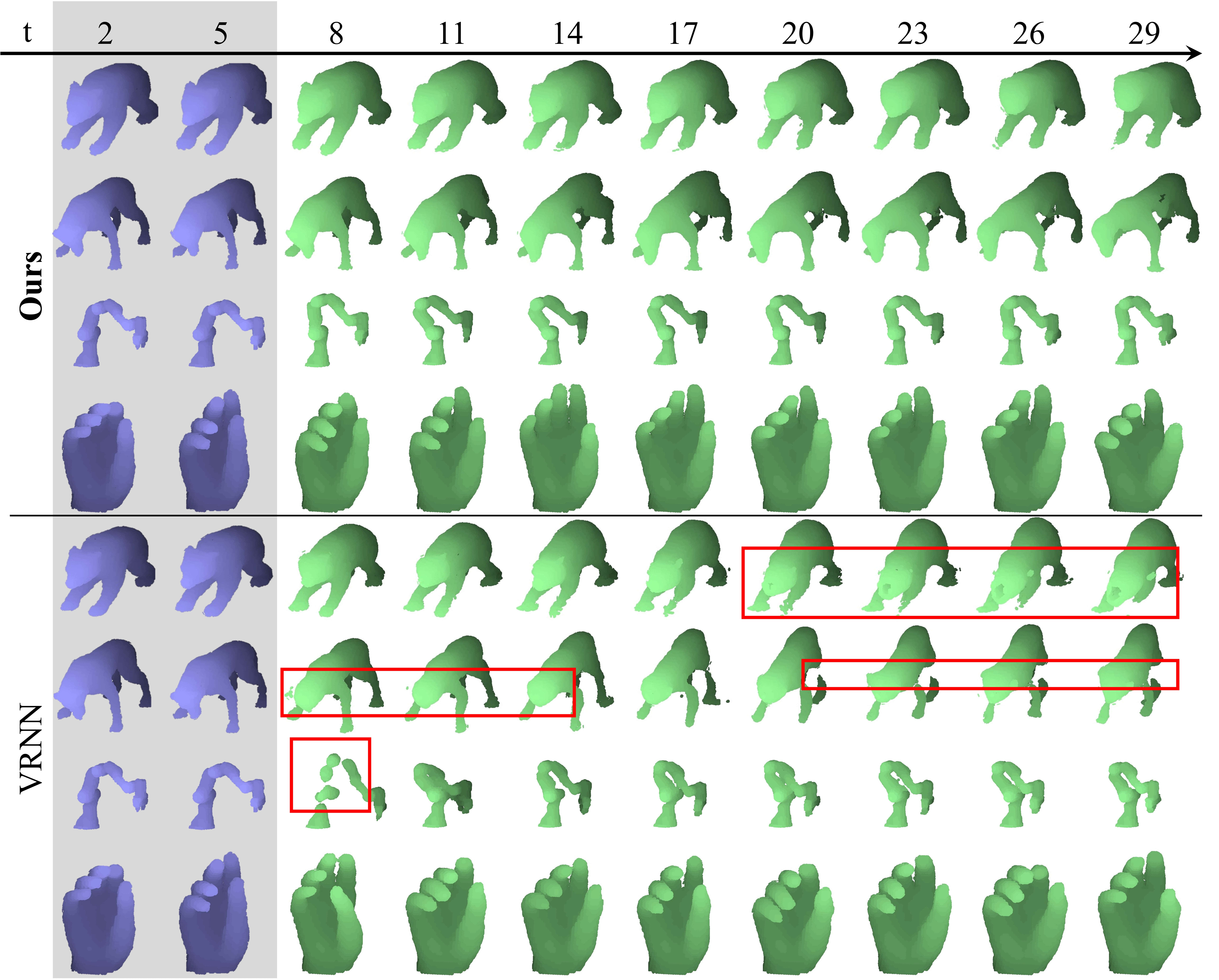} 
\caption{Additional samples for other categories: \textbf{Animals} for non-human animals, \textbf{Panda} for robot arms, and \textbf{HanCo} for human hands. Our model definitely outperforms in \textbf{Animals} dataset wherea VRNN also generates plausible sequence in \textbf{Panda} and \textbf{HanCo} dataset. Possible reason is that both \textbf{Panda} and \textbf{HanCo} dataset contains monotonous movements between neighboring frames so that objective  motion dynamics are easier to learn compared to \textbf{Animals} or \textbf{AIST++} dataset.}
\label{fig:additional_generation_others}
\end{figure}

\onecolumn
\subsection{D.3\quad Motion Interpolation}

\begin{figure}[hbt!]
\centering
\includegraphics[width=0.95\textwidth]{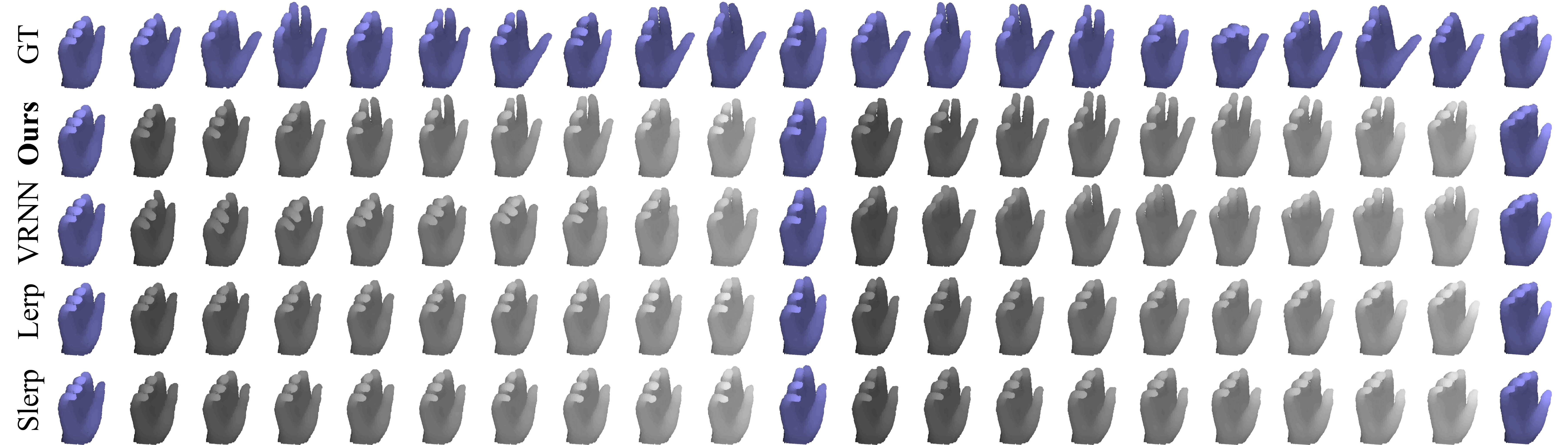} 
\caption{Additional motion interpolation results on \textbf{HanCo} dataset.}
\label{fig:additional_interploation_hanco}
\end{figure}

\begin{figure}[hbt!]
\centering
\includegraphics[width=0.95\textwidth]{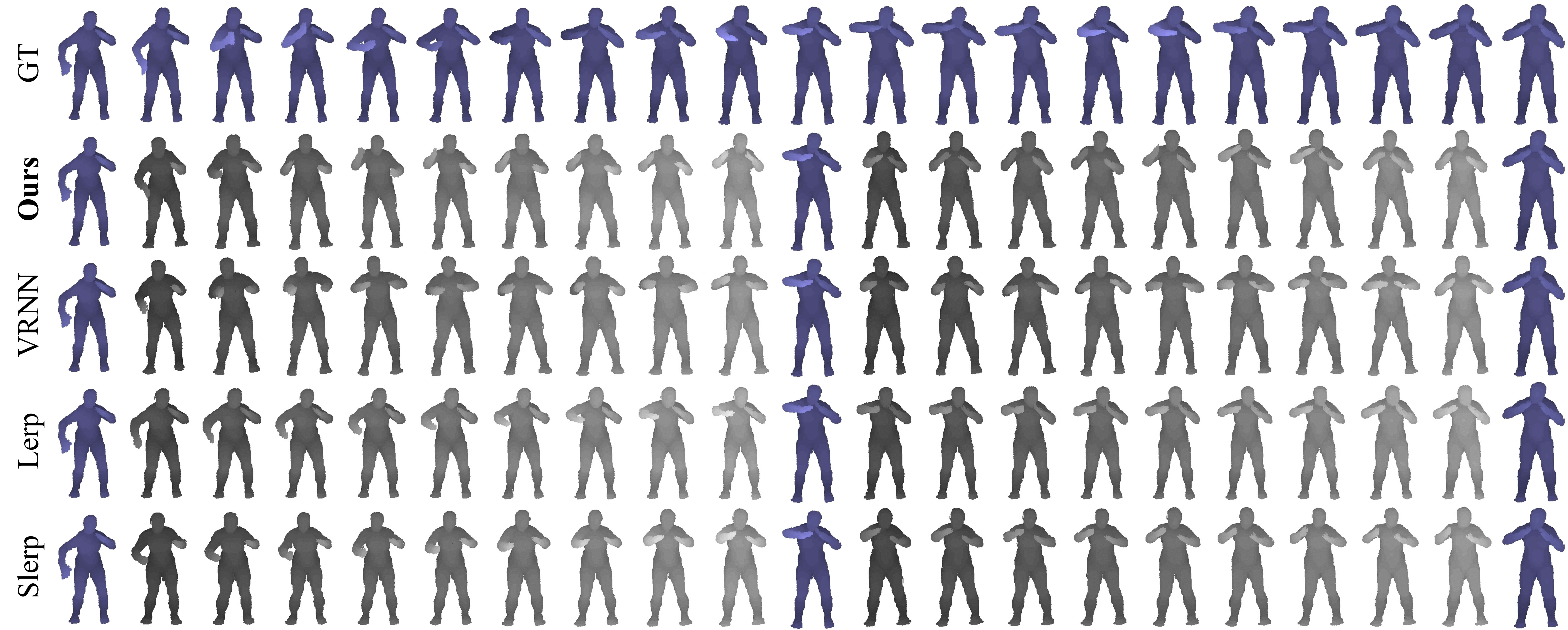} 
\caption{Additional motion interpolation results on \textbf{D-FAUST} dataset.}
\label{fig:additional_interploation_dfaust}
\end{figure}

\begin{table}[htb!]
\caption{Motion Chamfer Distance on motion interpolation results in \textbf{HanCo} dataset.}
\centering
\begin{small}
    \begin{tabular}{c|cccccccc}
         \toprule
        Models         & {SR = 3}       & {SR = 5}      & {SR = 10}     & {SR = 15}     & {SR = 20}     & {SR = 25}     & {SR = 30}  & Average\tabularnewline
        \midrule
        \textbf{Ours}  & \textbf{1.47} (0.0783)    & \textbf{1.98} (0.117)   &\textbf{2.69} (0.219)     &\textbf{2.96} (0.238)     &\textbf{3.09} (0.259)     & \textbf{3.08} (0.257)    & \textbf{3.20} (0.262) & \textbf{2.64}\\
        VRNN           & 1.48 (0.0839)    & 2.00 (0.135)   &2.80 (0.249)     &3.22 (0.263)     &3.64 (0.346)     & 3.69 (0.364)    & 4.02 (0.416) & 2.98\\
        Lerp           & 1.83 (0.225)    & 2.59 (0.208)   &3.98 (0.290)     &5.23 (0.354)     &6.52 (0.546)     & 7.33 (0.510)    & 7.98 (0.529) &5.07\\
        Slerp          & 1.71 (0.0921)    & 2.56 (0.127)   &4.04 (0.271)     &5.22 (0.335)     &6.45 (0.535)     & 7.15 (0.473)    & 7.74 (0.489) &4.52\\
        \bottomrule
    \end{tabular}
\end{small}

\label{tab:hanco_interpolation}
\end{table}

\begin{table}[htb!]
\caption{Motion Chamfer Distance on motion interpolation results in \textbf{D-FAUST} dataset.}
\centering
\begin{small}
    \begin{tabular}{c|cccccccc}
         \toprule
        Models         & {SR = 3}       & {SR = 5}      & {SR = 10}     & {SR = 15}     & {SR = 20}     & {SR = 25}     & {SR = 30}  & Average\tabularnewline
        \midrule
        \textbf{Ours}  & 1.61 (0.454)    & \textbf{1.92} (0.567)   &\textbf{2.50} (0.881)    &\textbf{3.08} (1.81)     &3.71 (2.62)              & \textbf{3.57} (2.18)    & \textbf{3.76} (2.27) & \textbf{2.88}\\
        VRNN           & \textbf{1.53} (0.448)    & 1.94 (0.618)   &2.63 (1.11)              &3.12 (1.42)              &\textbf{3.64} (1.65)     & 3.77 (2.00)    & 4.10 (2.19) & 2.96\\
        Lerp           & 1.84 (0.755)    & 2.34 (0.721)   &3.07 (0.859)     &3.74 (1.31)     &4.42 (1.40)     & 4.74 (1.38)    & 5.52 (1.65) & 3.67\\
        Slerp          & 1.84 (0.513)    & 2.29 (0.538)   &2.96 (0.790)     &3.74 (1.30)     &4.37 (1.48)     & 4.75 (1.59)    & 5.57 (1.79) & 3.65\\
        \bottomrule
    \end{tabular}
\end{small}

\label{tab:dfaust_interpolation}
\end{table}

\subsection{D.4\quad Motion Retargeting}

\paragraph{Retargeting process}
In this chapter, we additionally provide brief explanation on the motion retargeting process.
We first pretrain Neural Marionette with the \textbf{AIST++} dataset to learn the skeleton and motion dynamics of a human.

\begin{figure}[hbt!]
\centering
\includegraphics[width=0.90\textwidth]{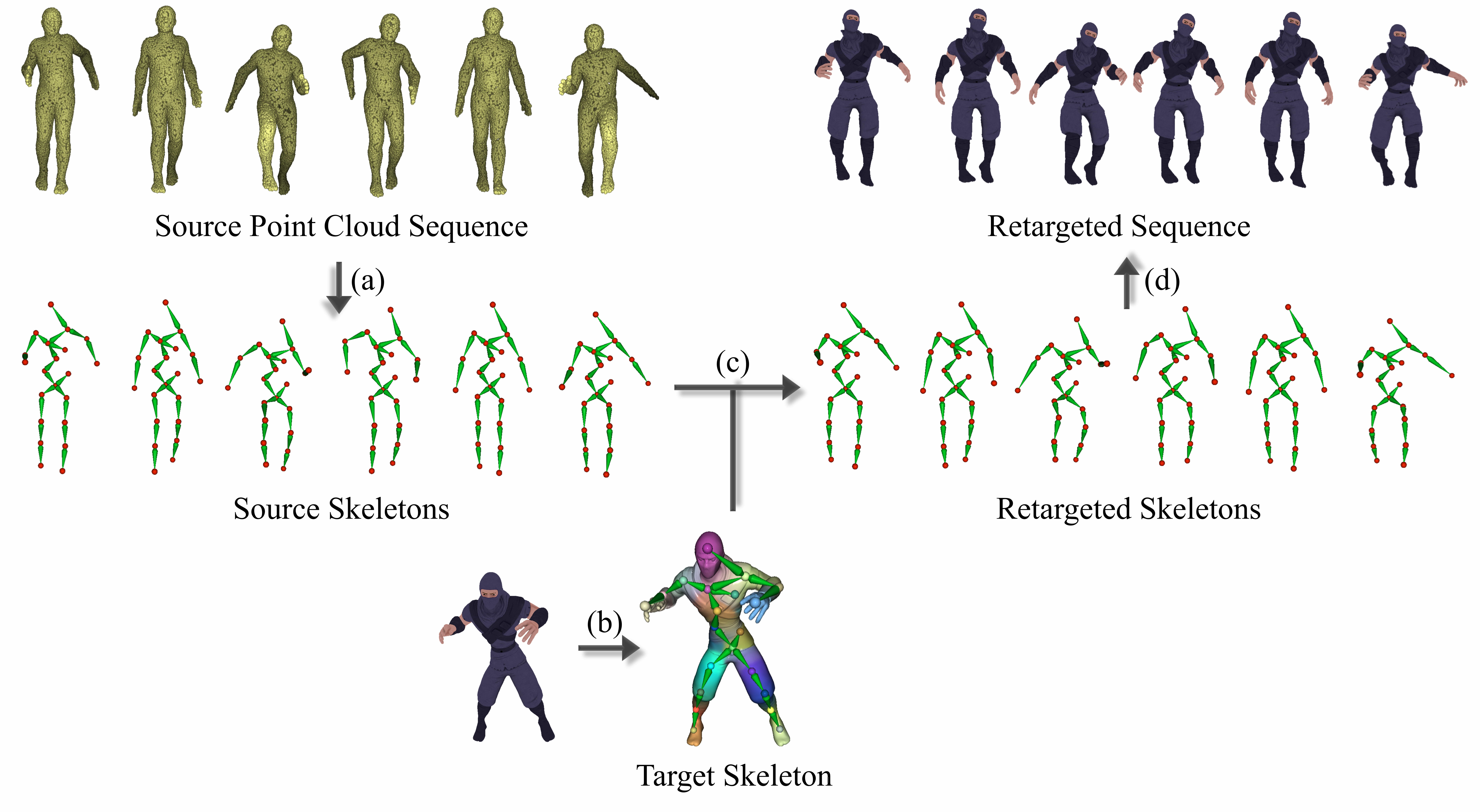} 
\caption{Process of motion retargeting.}
\label{fig:retargeting_explanation}
\end{figure}

Given the pre-trained neural networks, the retargeting process begins with extracting a stream of skeletons from the given point cloud sequence (Figure~\ref{fig:retargeting_explanation} (a)). 
Note that relative rotations are also estimated from the dynamics module.

At the same time, our pre-trained model also detects a skeleton from a single frame of the target shape, which are sampled points from the Mixamo humanoid meshes (Figure~\ref{fig:retargeting_explanation} (b)).
Surprisingly, we found that our model successfully extracted skeleton even though we did not use the mixamo data in the training.
Given the skeleton, we calculate the skin weights (Sec. A.4) of the target shape that we can deform the input shape with the retargeted skeletal motion.

Then we create a skeletal motion of the target shape from the source skeletal motion  (Figure~\ref{fig:retargeting_explanation} (c)).
Assuming that the source and target skeletons share the orientations of the offsets $\{\bar{d}_k\}_K$ of Eq.(14),
We simply apply the forward kinematics chain with the source rotations $\{R^{src}_k\}_K$ and target offsets $\{d^{tar}_k\}_K$ as
\begin{equation}
    \mu^{re}_{k,t} = \mu^{re}_{parent(k), t} + R^{src}_{k,t}d^{tar}_k~~\forall k,t~~\text{where}
    \label{eq:retarget_eq1}
\end{equation}
\begin{equation}
    d^{tar}_k = \bar{d}_k\|\mu^{tar}_k-\mu^{tar}_{parent(k)}\|_2~\forall k.
    \label{eq:retarget_eq2}
\end{equation}
Finally, we deform a target with the retargeted skeletons and skin weights following LBS process (Figure~\ref{fig:retargeting_explanation} (d)).

\begin{figure}[hbt!]
\centering
\includegraphics[width=0.95\textwidth]{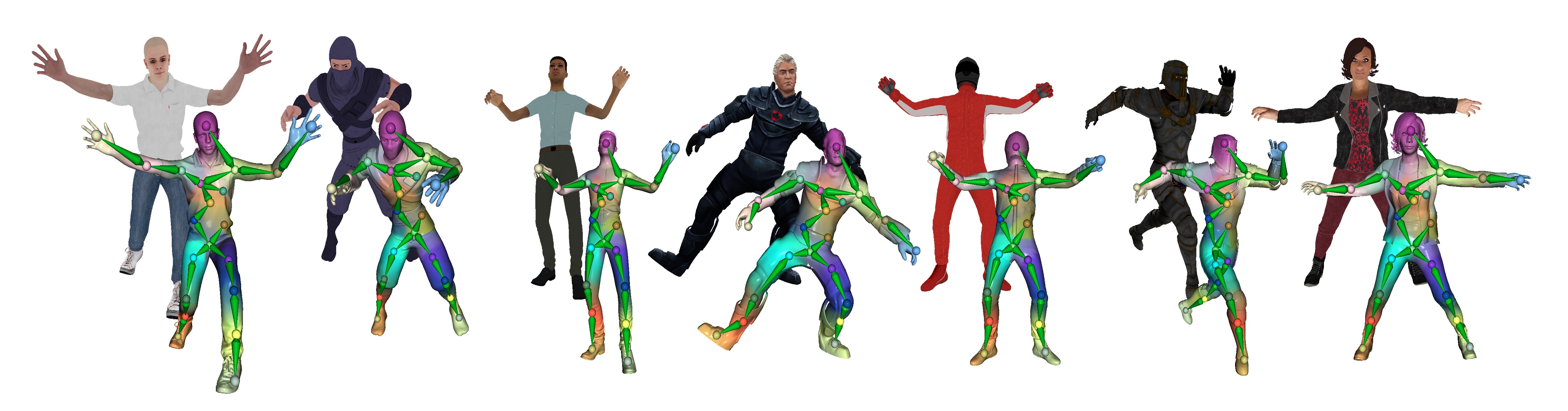} 
\caption{Additional rigging results of Mixamo data. Note that it is a zero-shot generalization as we never train our model with the given Mixamo humanoids.}
\label{fig:retargeting_target}
\end{figure}

\begin{figure}[hbt!]
\centering
\includegraphics[width=0.85\textwidth]{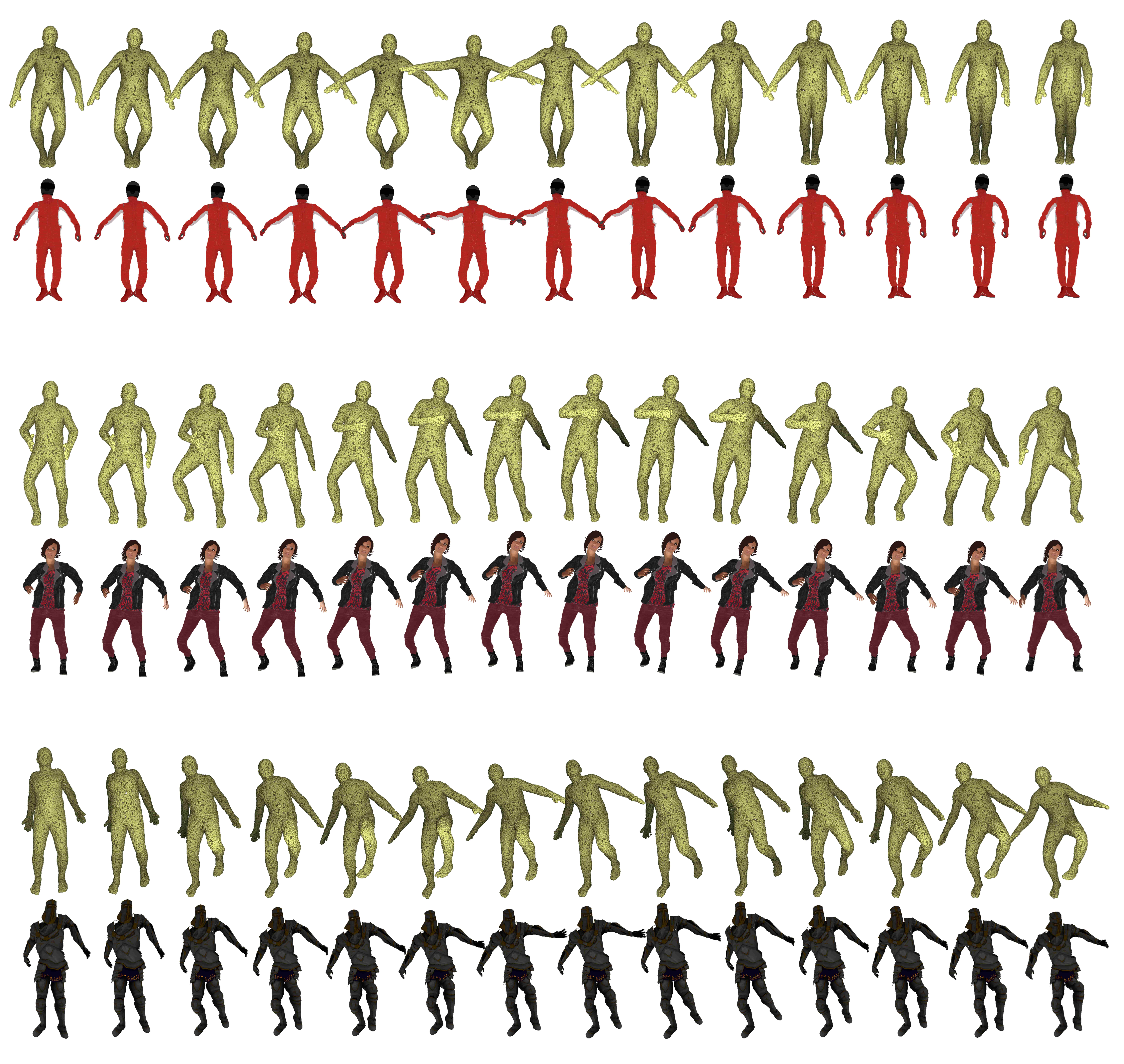} 
\caption{Additional results on motion retargeting.}
\label{fig:additional_retargeting}
\end{figure}

\end{document}